\def\year{2022}\relax
\documentclass[letterpaper]{article} 
\usepackage{aaai22}  
\usepackage{times}  
\usepackage{helvet}  
\usepackage{courier}  
\usepackage[hyphens]{url}  
\usepackage{graphicx} 
\usepackage{natbib}  
\usepackage{caption} 
\DeclareCaptionStyle{ruled}{labelfont=normalfont,labelsep=colon,strut=off} 
\usepackage{amsmath,amsfonts,amsthm,bm}
\usepackage{subfigure}
\usepackage{caption}
\usepackage[switch]{lineno}
\usepackage{multirow}
\usepackage{bbding}
\usepackage{bibentry}
\usepackage{algorithm}
\usepackage{algorithmic}
\usepackage{booktabs} 
\usepackage{newfloat}
\usepackage{listings}
\floatstyle{ruled}
\newfloat{listing}{tb}{lst}{}
\floatname{listing}{Listing}

\newtheorem{Proposition}{Proposition}
\newtheorem{Definition}{Definition}
\newtheorem{Example}{Example}
\title{Conditional Local Convolution for Spatio-temporal Meteorological Forecasting}

\usepackage{bibentry}
\newcommand{\printfnsymbol}[1]{
   \textsuperscript{\@*}
 }
\begin{document}
\author {
    Haitao Lin,\thanks{Equal Contributions.}\textsuperscript{\rm 1 \rm 3}
    Zhangyang Gao,\footnotemark[1]\textsuperscript{\rm 1}
    Yongjie Xu, \textsuperscript{\rm 1}
    Lirong Wu, \textsuperscript{\rm 1}
    Ling Li, \textsuperscript{\rm 2}
    Stan. Z. Li, \textsuperscript{\rm 1}
}
\affiliations {
    \textsuperscript{\rm 1} Center of Artificial Intelligence for Research and Innovation, Westlake University\\
    \textsuperscript{\rm 2} Eco-Environmental Research Laboratory, Westlake University
    \textsuperscript{\rm 3} Zhejiang University\\
    linhaitao, gaozhangyang, xuyongjie, wulirong, liling, stan.zq.li@westlake.edu.cn
}
\maketitle

\begin{abstract}
Spatio-temporal forecasting is challenging attributing to the high nonlinearity in temporal dynamics as well as complex location-characterized patterns in spatial domains, especially in fields like weather forecasting. Graph convolutions are usually used for modeling the spatial dependency in meteorology to handle the irregular distribution of sensors' spatial location.  In this work, a novel graph-based convolution for imitating the meteorological flows is proposed to capture the local spatial patterns. Based on the assumption of smoothness of location-characterized patterns, we propose conditional local convolution whose shared kernel on nodes' local space is approximated by feedforward networks, with local representations of coordinate obtained by horizon maps into cylindrical-tangent space as its input. The established united standard of local coordinate system preserves the orientation on geography. We further propose the distance and orientation scaling terms to reduce the impacts of irregular spatial distribution. The convolution is embedded in a Recurrent Neural Network architecture to model the temporal dynamics, leading to the Conditional Local Convolution Recurrent Network (CLCRN). Our model is evaluated on real-world weather benchmark datasets, achieving state-of-the-art performance with obvious improvements. We conduct further analysis on local pattern visualization, model's framework choice, advantages of horizon maps and etc.
The source code is available at \url{https://github.com/BIRD-TAO/CLCRN}.
\end{abstract}
\vspace{-0.3cm}
\section{1. Introduction}
In classical statistical learning, spatio-temporal forecasting is usually regarded as a multi-variate time series problem, and methods such as autoregressive integrated moving average (ARIMA) with its variants are proposed, but the stationary assumption is usually hard to satisfy. Recently, the rise of deep learning approaches has attracted lots of attention.
For example, in traffic flow forecasting, Graph Neural Networks (GNNs) are regarded as superior solutions to model spatial dependency by taking sensors as graph nodes \cite{yu2018STGCN,li2018diffusion}. Compared with great progress in traffic forecasting, works focusing on meteorology are scarce, while the need for weather forecasting is increasing dramatically \cite{shi2015convolutional, sonderby2020metnet}. In this work, our research attention is paid to spatio-temporal meteorological forecasting tasks.

The task is challenging due to two main difficulties. First, the irregular sampling of meteorological signals usually disables the classical Convolutional Neural Networks (CNNs) which work well on regular mesh grid signals on Euclidean domains such as 2-D planar images. Signals are usually acquired from irregularly distributed sensors, and the manifolds from which signals are sampled are usually non-planar. For example, sensors detecting temperature are located unevenly on land and ocean, which are not fixed on structured mesh grids, and meteorological data are often spherical signals rather than planar ones. Second, the high temporal and spatial dependency makes it hard to model the dynamics. For instance, different landforms show totally distinct wind flow or temperature transferring patterns in weather forecasting tasks and extreme climate incidents like El Nino \cite{elnino} often cause non-stationarity for prediction. 

GNNs yield effective and efficient performance for irregularly spatio-temporal forecasting, enabling to update node representations by aggregating messages from their neighbors, the process of which can be analogized to heat or wind flow from localized areas on the earth surface. As discussed, the meteorological flow may demonstrate totally variant patterns in different local regions. Inspired by the analogy and location-characterized patterns, we aim to 
establish a graph convolution kernel, which varies in localized regions to approximate and imitate the true local meteorological patterns.

Therefore, we propose our \textbf{conditional local kernel}. We embed it in a graph-convolution-based recurrent network, for spatio-temporal meteorological forecasting. The convolution is performed on the \textbf{local space} of each node, which is constructed considering both distance between nodes and their relative orientation, with the kernel proposed mainly based on the assumption: \textbf{smoothness of location-characterized patterns}.  In summary, our contributions are:
\begin{itemize}
    \item Proposing a location-characterized kernel to capture and imitate the meteorological local spatial patterns in its message-passing process;
    \item Establishing the spatio-temporal model with the proposed graph convolution which achieves state-of-the-art performance in weather forecasting tasks;
    \item Conducting further analysis on learned local pattern visualization, framework choice, local space and map choice and ablation.
\end{itemize}

\section{2. Related Work}
\label{sec:relatedwork}
\paragraph{Spherical signal processing.} The spatial signals in meteorology are usually projected on a sphere, e.g. earth surface. Different from regular mesh grid samples on plane, sampling space of spherical signals demonstrates different manifold properties, which needs a specially designed convolution to capture the spatial pattern, such as multi-view-projection-based 2-D CNN \cite{coor2018spherenet}, 3-D mesh-based convolution \cite{jiang2019spherical} and graph-based spherical CNN \cite{PERRAUDIN2019130}. Further proposed mesh-based convolutions have remarkable hard-baked properties, such as in \cite{cohen2018spherical} and \cite{esteves2018learning} with equivariance, in spite of high computational cost and requirements of mesh grid data. For signals on irregular spatial distribution, graph-based neural networks are usually employed, representing the nodes on sphere as nodes of the established graph, with fast implementation and good performance \cite{Defferrard2020DeepSphere}. Our method also processes spherical signals with graph-based methods, and harnesses properties of sphere manifold, to model the local patterns in meteorology. 

\paragraph{Spatio-temporal graph neural networks.} Graph neural networks perform convolution based on the graph structure, and yield effective representations by aggregating or diffusing messages from or to neighborhoods \citep{niepert2016learning, atwood2016diffusionconvolutional,kipf2017semisupervised} or filter different frequency based on graph Laplacian \cite{bruna2014spectral, defferrard2017convolutional}.  After the rise of GNNs \cite{gilmer2017neural, veli2018graph, wu2021selfsupervised}, spatio-temporal forecasting models are mostly graph-based neural networks thanks to their ability to learn representations of spatial irregular distributed signals, such as STGCN \cite{yu2018STGCN} convoluting spatial signals by using spectral filters for traffic forecasting, and DCRNN \cite{li2018diffusion} achieving tremendous improvements on the same task by employing diffusion convolution on graphs. Despite the inability of previous methods to adaptively model location-characterized patterns, graph attention \cite{Guo2019ASTGCN} is employed in spatio-temporal model to learn the adjacency relation among traffic sensors, and an adaptive graph recurrent model \cite{bai2020adaptive} is proposed to optimize different local patterns according to higher-level node representations. Differing from the previous, the adaptively learned local patterns in our method are hard-baked with property of local smoothness of location-characterized patterns, thus capable of capturing the meteorological flowing process. 
\section{3. Background}
\paragraph{Problem setting.} Given $N$ correlated signals located on the sphere manifold $S^2$ at time  $t$, we can represent the signals as a (directed) graph $\mathcal{G} = (\mathcal{V}, \mathcal{E}, \bm{A})$, where $\mathcal{V}$ is a node set with $\mathcal{V} = \{\bm{\mathrm{x}}_i^S = (x_{i,1}, x_{i,2}, x_{i,3})\in S^2: i=1,2,\ldots,N\}$ meaning that it records the position of $N$ nodes, which satisfies $||\bm{\mathrm{x}}_i^S||_2 = 1$. We denote positions of nodes in Euclidean space by $\bm{\mathrm{x}}^E$, and in sphere as $\bm{\mathrm{x}}^S$. $\mathcal{E}$ is a set of edges and $\bm{A} \in \mathbb{R}^{N \times N}$ is the adjacency matrix which can be asymmetric. The signals observed at time $t$ of the nodes on $\mathcal{G}$ are denoted by $\bm{F}^{(t)} \in \mathbb{R}^{N\times D}$. For the forecasting tasks, our goal is to learn a function $P(\cdot)$ for approximating the true mapping of historical $T'$ observed signals to the future $T$ signals, that is
\begin{align}
    [\bm{F}^{(t-T')}, \ldots, \bm{F}^{(t)};\mathcal{G} ] \overset{P}{\longrightarrow} [\bm{F}^{(t+1)}, \ldots, \bm{F}^{(t+T)};\mathcal{G}].
\end{align}
In the paper, for meteorological datasets which do not provide the adjacency matrix, we construct it by K-nearest neighbors algorithm based on induced spherical distance of their spatial location which will be discussed later. 
\paragraph{Graph convolution neural networks.} For notation simplicity, we omit the supper-script $(t)$ when discussing spatial dependency. Denote the set of neighbors of center node $i$ by  $\mathcal{N}(i) = \{j:(i,j) \in \mathcal{E}\}$, and note that $(i,i) \in \mathcal{N}(i)$. In Graph Neural Networks, $(W^l, \bm{b}^l)$ is the weights and bias parameters for layer $l$, and $\sigma(\cdot)$ is a non-lieanr activation function. The message-passing rule concludes that at layer $l$, representation of node $i$ updates as
\begin{align}
     \bm{y}^{l}_i &= \sum_{j\in \mathcal{N}(i)}\omega_{i,j}\bm{h}^{l-1}_j; \label{eq:messgagepass}\\
    \bm{h}^{l}_i &= \sigma(\bm{y}^{l}_i\bm{W}^l + \bm{b}^l) ,
\end{align}
where $\bm{h}^{l}_i$ is the representation of node $i$ after $l$-th layer, with $\bm{h}^{0}_i = \bm{F}_i$, which is the observed graph signals on node $i$.  Denote the neighborhood coordinate set of center node $i$ by $\mathcal{V}(i) = \{\bm{\mathrm{x}}^S_j: j\in \mathcal{N}(i)\}$, and then Eq.~\ref{eq:messgagepass} represents aggregation of messages from neighbors, which can also be regarded as the convolution operation on graph, which can be written as 
{\small
\begin{align}
    (\Omega \star_{\mathcal{N}(i)} \bm{H}^{l-1})(\bm{\mathrm{x}}_j^S) &= \sum_{\bm{\mathrm{x}}^S_j\in \mathcal{V}(i)} \Omega(\bm{\mathrm{x}}^S_j; \bm{\mathrm{x}}^S_i)\bm{H}^{l-1}(\bm{\mathrm{x}}^S_j), \label{eq:graphconv}
\end{align}
}%
where $\star_{\mathcal{N}(i)}$ means convolution on the $i$-th node's neighborhood, $\Omega: S^2 \times S^2 \rightarrow \mathbb{R}$ is the convolution kernel, such that $\Omega(\bm{\mathrm{x}}^S_j, \bm{\mathrm{x}}^S_i) = \omega_{i,j}$, and $\bm{H}^{l-1}$ is a function mapping each point on sphere to its feature vector in $l$-th representation space. 
\begin{Example}\rm The convolutional kernel used in DCRNN \cite{li2018diffusion} is
\begin{align}
\Omega(\bm{\mathrm{x}}^S_j, \bm{\mathrm{x}}^S_i) = 
  &\exp(-d^2(\bm{\mathrm{x}}^S_i, \bm{\mathrm{x}}^S_j)/\tau),
\end{align}
where $d(\cdot,\cdot)$ is the distance between the two nodes, and $\tau$ is a hyper-parameter to control the smoothness of the kernel.
\end{Example}
To imitate the meteorological patterns, the value of convolution kernel should be large for neighbors having great meteorological impacts on centers. For example, If there exists heat flows from the south-east to the north-west, the kernel should give more weights to the nodes from the south-east when aggregating messages from neighbors.
Using slight abuse of terminology, we consider the convolution kernels are equivalent to meteorological patterns in local regions. 
\paragraph{Sphere manifold.} The signals are located on the earth surface, which is regarded as a sphere, and thus we introduce the notation of sphere manifold to further develop our convolution method. The $M$-D sphere manifold is denoted by $S^M = \{\bm{\mathrm{x}}^S = (x_1, x_2,\ldots, x_{M+1})\in \mathbb{R}^{M+1}:||\bm{\mathrm{x}}^S|| = 1\}$. 
The convolution is usually operated on a plane, so we introduce the \textbf{local space}, the $M$-D Euclidean space, as the convolution domains.
\begin{Definition}
Define the local space centered at point $\bm{\mathrm{x}}$ as some Euclidean space denoted by $\mathcal{L}_{\bm{\mathrm{x}}}S^M$, with $\bm{\mathrm{x}} \in \mathcal{L}_{\bm{\mathrm{x}}}S^M$, which is homeomorphic to the local region centered at $\bm{\mathrm{x}}$. \rm{(Formal definition see Appendix A2.)}
\end{Definition}

\begin{Example}\rm
The tangent space centered at point $\bm{\mathrm{x}}$ is an example of local space, denoted by $\mathcal{T}_{\bm{\mathrm{x}}}S^M =  \{\bm{\mathrm{v}}\in \mathbb{R}^{M+1}: <\bm{\mathrm{x}}, \bm{\mathrm{v}}> = 0\}$, where $<\cdot,\cdot>$ is the Euclidean inner product.
\end{Example} 

The \textbf{geodesics and induced distance} on sphere are important to both defining the neighborhood of a node, as well as identifying the message-passing patterns. Intuitively, the greater is the distance from one node to another, the fewer messages should be aggregated from the node into another in graph convolution. 
\begin{Proposition}
Let $\bm{\mathrm{x}} \in S^{M}$, and $\bm{\mathrm{u}} \in \mathcal{T}_{\bm{\mathrm{x}}}S^M$ be unit-speed. The unit-speed geodesics is $\gamma_{\bm{\mathrm{x}}\rightarrow\bm{\mathrm{u}}}(t) = \bm{\mathrm{x}} \cos t + \bm{\mathrm{u}} \sin t$, with $\gamma_{\bm{\mathrm{x}}\rightarrow\bm{\mathrm{u}}}(0) = \bm{\mathrm{x}}$ and $\dot \gamma_{\bm{\mathrm{x}}\rightarrow\bm{\mathrm{u}}}(0) = \bm{\mathrm{u}}$. The intrinsic shortest distance function between two points $\bm{\mathrm{x}}, \bm{\mathrm{y}} \in S^M$ is
\begin{align}
     d_{S^M}(\bm{\mathrm{x}}, \bm{\mathrm{y}}) = \arccos(<\bm{\mathrm{x}},\bm{\mathrm{y}}>).
\end{align}
\end{Proposition}
The distance function is usually called great-circle distance on sphere. In practice, the K-nearest neighbors algorithm to construct the graph structure is conducted based on spherical distance.

On the establishment of local space of each center node, an \textbf{isometric map} $\mathcal{M}_{\bm{\mathrm{x}}}(\cdot): S^M \rightarrow \mathcal{L}_{\bm{\mathrm{x}}}S^M$ satisfying that $||\mathcal{M}_{\bm{\mathrm{x}}}(\bm{\mathrm{y}})|| = d_{S^M}(\bm{\mathrm{x}},\bm{\mathrm{y}})$ can be used to map neighbor nodes on sphere into the local space. 

\begin{Example}\rm \label{ex:logmap}
Logarithmic map is usually used to map the neighbor node $\bm{\mathrm{x}}_j\in \mathcal{V}(i)$ on sphere isometrically into $\mathcal{T}_{\bm{\mathrm{x}}_i}S^M$, which reads
\begin{align*}
    \log_{\bm{\mathrm{x}}_i}(\bm{\mathrm{x}}_j) &= d_{S^M}(\bm{\mathrm{x}}_i, \bm{\mathrm{x}}_j)\frac{P_{{\bm{\mathrm{x}}_i}}(\bm{\mathrm{x}}_j - \bm{\mathrm{x}}_i)}{||P_{{\bm{\mathrm{x}}_i}}(\bm{\mathrm{x}}_j - \bm{\mathrm{x}}_i)||} ,
\end{align*}
where $P_{{\bm{\mathrm{x}}_i}}(\bm{\mathrm{x}}) = \frac{\bm{\mathrm{x}}}{||\bm{\mathrm{x}}||} - <\frac{\bm{\mathrm{x}}_i}{||\bm{\mathrm{x}}_i||},\frac{\bm{\mathrm{x}}}{||\bm{\mathrm{x}}||}>\frac{\bm{\mathrm{x}}_i}{||\bm{\mathrm{x}}_i||}$ is the normalized projection operator.
\end{Example}

After the neighbors of $\bm{\mathrm{x}}_i$ are mapped into local space of the center nodes through the isometric maps, which reads $\bm{\mathrm{v}}_j = \mathcal{M}_{\bm{\mathrm{x}}_i}(\bm{\mathrm{x}}_j)$, the \textbf{local coordinate system} of each center node is set up, through a transform mapping $\Pi_{\bm{\mathrm{x}}_i}(\bm{\mathrm{v}}_j) = \bm{\mathrm{x}}_j^{i'}$ for each $\bm{\mathrm{x}}_j \in \mathcal{V}(i)$. We call $\bm{\mathrm{x}}_j^{i'}$ the relative position of $\bm{\mathrm{v}}_j$ in the local coordinate system of the local space centered at  $\bm{\mathrm{x}}_i$. As $\bm{\mathrm{x}}_j^{i'}$ is always in the local space which is Euclidean, the supper-script $E$ is omitted.  The mapping  $\Pi_{\bm{\mathrm{x}}_i}(\cdot)$ can be determined by $M$ orthogonal basis chosen in the local coordinate system, i.e. $\{\bm{\xi}^1, \bm{\xi}^2, \ldots, \bm{\xi}^M\}$, which will be discussed later in $S^2$ scenario for meteorological application.
\section{4. Proposed Method}
\subsection{4.1. Local convolution on sphere}
Given a center node ${\bm{\mathrm{x}}_i^E} \in \mathbb{R}^2$, from the perspective of defined graph convolution in Eq.~\ref{eq:graphconv}, the convolution on planar mesh grids such as pixels on images is written as
{\small
\begin{align}
    (\Omega \star_{\mathcal{V}{(i)}} \bm{H})(\bm{\mathrm{x}}_i^E) &= \sum_{\bm{\mathrm{x}}^E} \Omega(\bm{\mathrm{x}}^E; \bm{\mathrm{x}}^E_i) \bm{H}(\bm{\mathrm{x}}^E)\delta_{\mathcal{V}(i)}(\bm{\mathrm{x}}^E)\notag\\ 
    &=\sum_{\bm{\mathrm{x}}^E} \chi(\bm{\mathrm{x}}^E_i - \bm{\mathrm{x}}^E) \bm{H}(\bm{\mathrm{x}}^E)\delta_{\mathcal{V}(i)}(\bm{\mathrm{x}}^E),
\end{align}
}%
where $\delta_{\mathcal{A}}(\bm{\mathrm{x}}) = 1$ if $\bm{\mathrm{x}} \in \mathcal{A}$, else $0$.  In terms of convolution on 2-D images, $\mathcal{V}(i) = \{\bm{\mathrm{x}}^E :\bm{\mathrm{x}}^E - \bm{\mathrm{x}}_i^E \in \mathbb{Z}^2\cap ([-k_1, k_1] \times [-k_2, k_2])\}$.  $k_1 > 0$ and $k_2 >0$ are the convolution views to restrict how far away pixels are included in the neighborhood along the width-axis and length-axis respectively. When $k_1, k_2 < +\infty$, the neighborhood set is limited, and thus the convolution is defined as \textbf{local}, conducted on each node's local space, with \textbf{local convolution kernel} $\chi(\cdot)$ .


To extend the local convolution on generalized manifolds, we conclude that the local space of $\bm{\mathrm{x}}^E_i$ is $\mathcal{L}_{\bm{\mathrm{x}}_i^E}\mathbb{R}^2 = \{\bm{\mathrm{x}}^E :\bm{\mathrm{x}}^E - \bm{\mathrm{x}}_i^E \in[-k_1, k_1] \times [-k_2, k_2]\}$, so that the isometric map satisfies  $\bm{\mathrm{v}}^E = \mathcal{M}_{\bm{\mathrm{x}}_i}(\bm{\mathrm{x}}^E) =  \bm{\mathrm{x}}^E - \bm{\mathrm{x}}^E_i $. $\{-\bm{e}_{x}, -\bm{e}_y\}$ with $\bm{e}_{x}=(1,0)$ and $\bm{e}_{y}=(0,1)$ is the orthogonal basis in local coordinate system of the local space. In conclusion, 
\begin{align}
   \bm{\mathrm{x}}^{i'}=\Pi_{\bm{\mathrm{x}}_i^E}(\bm{\mathrm{v}}^E) =  -\bm{\mathrm{v}}^E = \bm{\mathrm{x}}_i^E - \bm{\mathrm{x}}^E .
\end{align}
In this way, we obtain the local convolution on 2-D Euclidean plane, which reads
{\small
\begin{align}
    (\Omega \star_{\mathcal{V}{(i)}} \bm{H})(\bm{\mathrm{x}}_i^E) &= \sum_{\bm{\mathrm{x}}^E} \chi(\bm{\mathrm{x}}^{i'}) \bm{H}(\bm{\mathrm{x}}^E)\delta_{\mathcal{V}(i)}(\bm{\mathrm{x}}^E).
\end{align}}%
In analogy to this, the local convolution on 2-D spherical the center node $\bm{\mathrm{x}}^S_i$ is defined similarly:
{\small\begin{align}
    (\Omega \star_{\mathcal{V}{(i)}} \bm{H}) (\bm{\mathrm{x}}_i^E)
    &=\sum_{\bm{\mathrm{x}}^S} \chi(\bm{\mathrm{x}}^{i'}) \bm{H}(\bm{\mathrm{x}}^S)\delta_{\mathcal{V}(i)}(\bm{\mathrm{x}}^S),
\end{align}
}%
where $\mathcal{V}(i)$ is given by the graph structure, nodes in which can be mapped into $\bm{\mathrm{x}}_i^S$'s local space. And $\bm{\mathrm{x}}^{i'}$ is obtained by:
\begin{align}
     \bm{\mathrm{x}}^{i'}=\Pi_{\bm{\mathrm{x}}_i^S}(\bm{\mathrm{v}}^S) =  \Pi_{\bm{\mathrm{x}}_i^S}(\mathcal{M}_{\bm{\mathrm{x}}_i^S}(\bm{\mathrm{x}}^S)) .
\end{align}
Following parts are organized to discuss how to elaborate 
\begin{itemize}
    \item $\mathcal{M}_{\bm{\mathrm{x}}_i^S}(\cdot)$ and $\Pi_{\bm{\mathrm{x}}_i^S}(\cdot)$, isometry mapping neighbors into some local space of ${\bm{\mathrm{x}}_i^S}$ and choice of orthogonal basis in local coordinate system.
    \item $\chi(\cdot)$, the formulation of convolution kernel to approximate and imitate the meteorological patterns.
\end{itemize}


\subsection{4.2. Orientation-preserving local regions}
In the following parts, all the nodes are located on sphere, so the supper-script $S$ is omitted. We choose what we define as \textbf{cylindrical-tangent space} and \textbf{horizon maps} (Fig. 1(a).) to construct local spaces and to map neighbors into them.  
\begin{Definition}
For $\bm{\mathrm{x}}_i  \in S^2$, the cylindrical-tangent space centered at $\bm{\mathrm{x}}_i$ reads
\begin{align*}
     \mathcal{C}_{\bm{\mathrm{x}}_i} S^2 = \{\bm{\mathrm{v}}\in \mathbb{R}^3:<\bm{\mathrm{v}}^-, \bm{\mathrm{x}}_i^-> = 0\},
\end{align*}
where $\bm{\mathrm{x}}^- = (x_1, x_2)$, taking the first two coordinates of vectors in $\mathbb{R}^3$.
\end{Definition}
\begin{Proposition}
Similar to logarithmic map, the horizon map $\mathcal{H}_{\bm{\mathrm{x}}_i}(\cdot)$ is used to map the neighbor node $\bm{\mathrm{x}}_j\in \mathcal{V}(i)$ isometrically into $\mathcal{C}_{\bm{\mathrm{x}}_i} S^2$, which reads
\begin{align*}
    \mathcal{H}_{\bm{\mathrm{x}}_i}(\bm{\mathrm{x}}_j) = d_{S^2}(\bm{\mathrm{x}}_i, \bm{\mathrm{x}}_j)\frac{[P_{{\bm{\mathrm{x}}_i^-}}(\bm{\mathrm{x}}_j^- - \bm{\mathrm{x}}_i^-), x_{j,3} - x_{i,3}]}{||[P_{{\bm{\mathrm{x}}_i^-}}(\bm{\mathrm{x}}_j^- - \bm{\mathrm{x}}_i^-), x_{j,3} - x_{i,3}]||},
\end{align*}
where $[\cdot,\cdot]$ is the concatenation of vectors/scalars.
\end{Proposition}
\begin{figure*}[ht]\vspace{-0.6cm}
\centering
        \subfigure[local space and isometric maps.]{ 
			\includegraphics[height=0.27\linewidth]{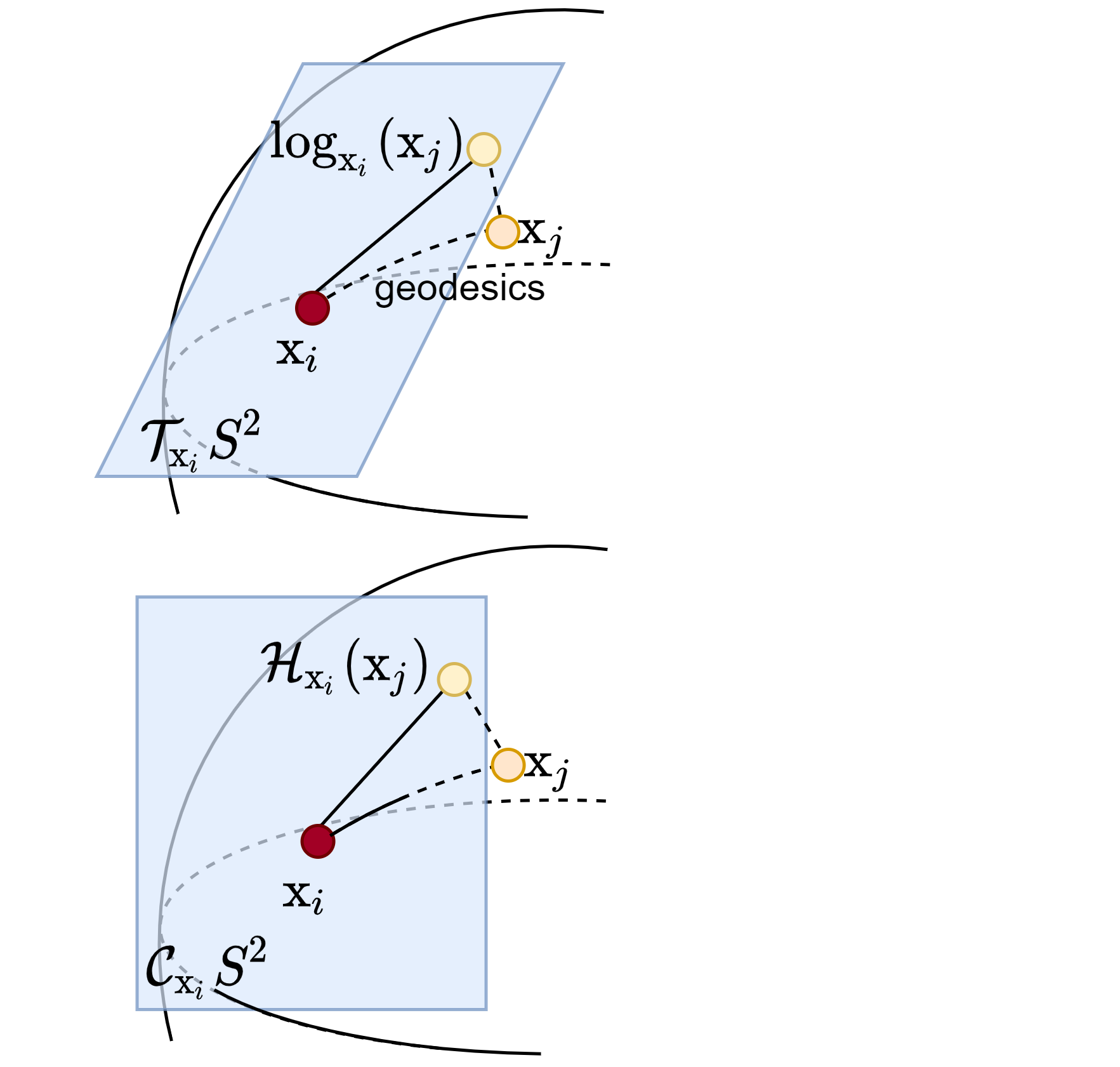}}\hspace{0mm}
		\subfigure[Example of unified standard of basis]{ \label{fig:exampleunify}
			\includegraphics[width=0.31\linewidth]{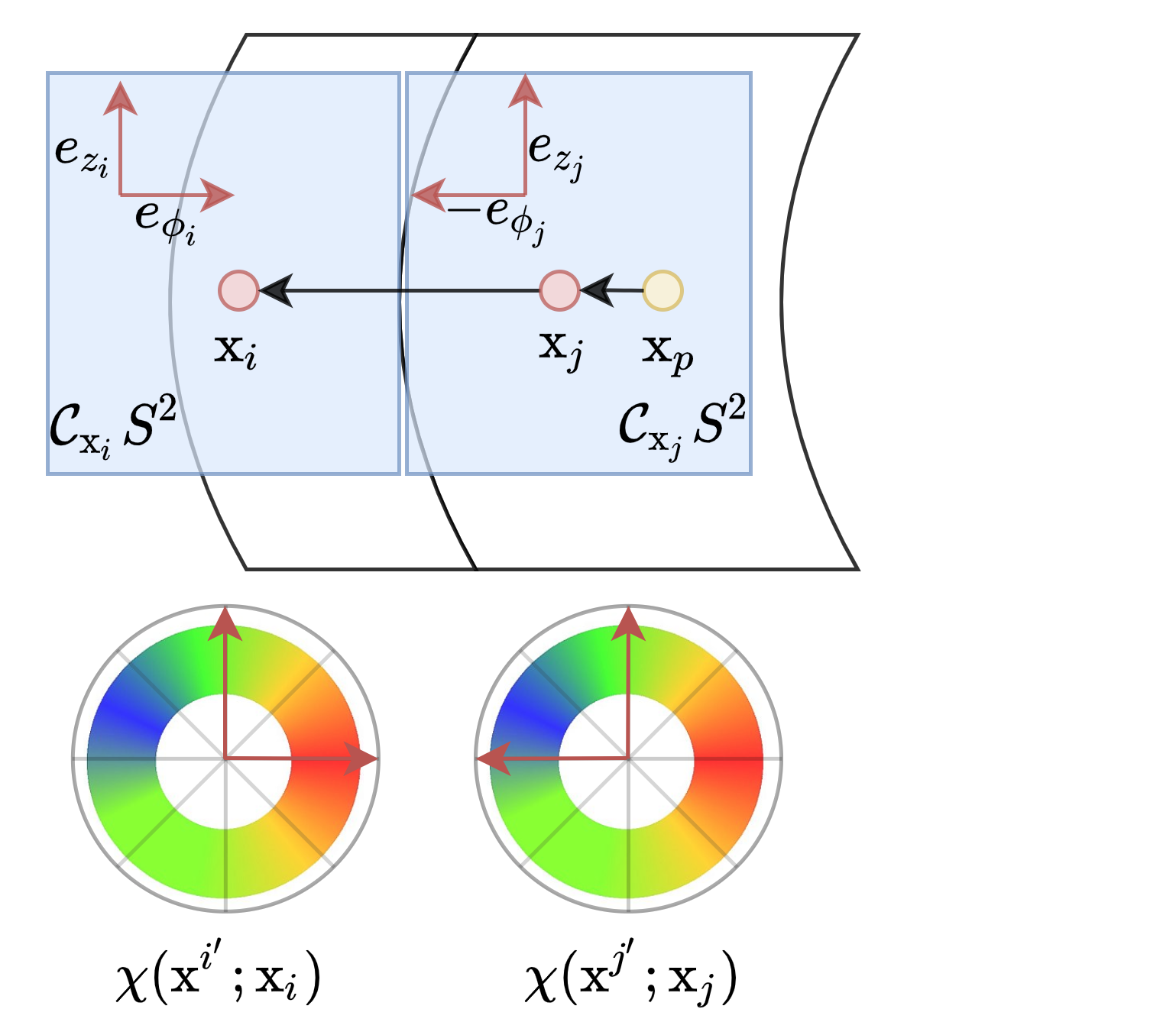}}\hspace{2mm}
		\subfigure[Example of reweighting angle scale]{ \label{fig:examplereweight}
			\includegraphics[width=0.31\linewidth]{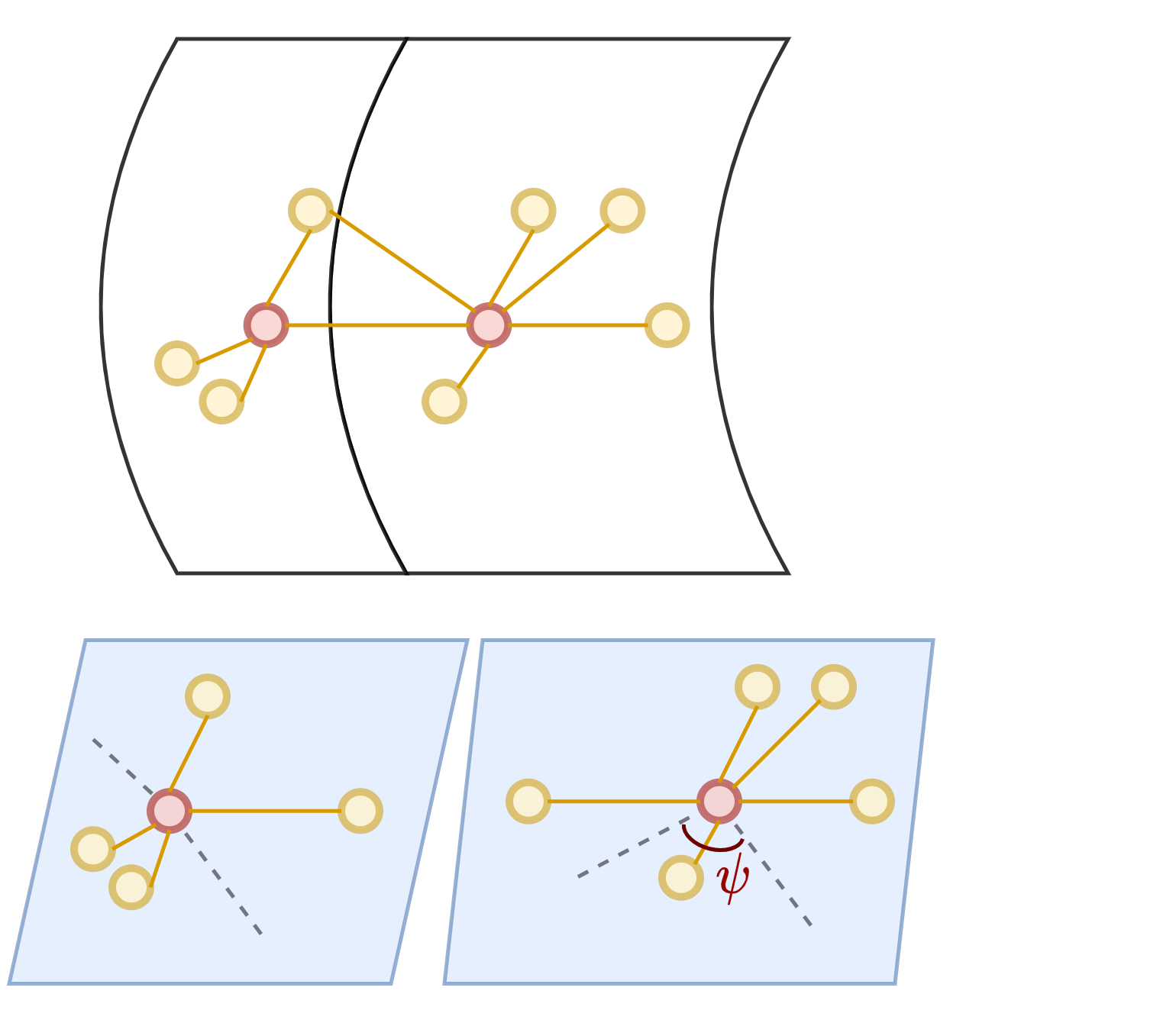}} \vspace{-0.3cm}
\caption{(a) shows tangent space with logarithmic maps in the top and cylindrical-tangent space with horizon maps in the below. (b) shows the necessity of the unified standard for choice of the basis. $\bm{\mathrm{x}}_p$ from the east affects both $\bm{\mathrm{x}}_i$ and $\bm{\mathrm{x}}_j$ a lot, with the corresponding local patterns in two center nodes shown in the heatmaps. However, if the basis is not unified as given in the example, the smoothness of local convolution kernel will be compromised. (c) shows the motivation of reweighting the angle scale. The angle scale $\frac{\psi}{2\pi}$ is given to balance neighbors' uneven contributions to the centers resulted from irregular distribution.}\vspace{-0.4cm}
\end{figure*}
The reason for choosing cylindrical-tangent space with horizon maps rather than tangent space with logarithmic maps is that the former one preserves the relative orientation on geographic graticules on the earth surface, which has explicitly geophysical meaning in meteorology. The logarithmic maps distort the relative position in orientation on graticules. For a node in the northern hemisphere, a neighbor locate in the east of it will locate in the north-east on its tangent plane after logarithmic map. In comparison, the defined cylindrical-tangent space preserves both relative orientation on graticules and spherical distance after mapping. Detailed proofs are provided Appendix B1 and empirical comparisons are given in Experiments 5.4.

As discussed, the cylindrical-tangent space is Euclidean, so in $S^2$, the transform  $\Pi_{\bm{\mathrm{x}}_i}(\cdot)$ can be determined by two orthogonal bases which are not unique. 
 Since our method is mainly implemented in spherical meteorological signals, we choose $\{\bm{e}_{\phi},\bm{e}_z\}$ as the two orthogonal bases in every local coordinate system of the cylindrical-tangent plane, in order to permit every local space to share the consistent South and North poles and preserve the relative position. For $\bm{\mathrm{x}}_i = (x_{i,1}, x_{i,2}, x_{i,3})$ and $\bm{\mathrm{v}} \in \mathcal{C}_{\bm{\mathrm{x}}_i}S^2$, let $\phi_i = \arctan{(x_{i,2}/x_{i,1})}$, and $\bm{\mathrm{x}}^{i'} = \Pi_{\bm{\mathrm{x}}_i}(\bm{\mathrm{v}}) = (\theta^{i'}, z^{i'})$ which is obtained by 
\begin{align}
    \phi^{i'} = <\bm{\mathrm{v}}, \bm{e}_{\phi_i}>;\quad 
    z^{i'}  = <\bm{\mathrm{v}}, \bm{e}_{z_i}>,
\end{align}
which is the latitude and longitude on the sphere, and 
\begin{align}
 \label{eq:orthobasis}
\bm{e}_{\phi_i} = (-\sin\phi_i, \cos\phi_i, 0); \quad
\bm{e}_{z_i} = (0, 0, 1).
\end{align}
The discussed maps and transforms cannot be applied for the South and North pole. We discuss it in Appendix B2.
\subsection{4.3. Conditional local convolution}
Now we introduce the conditional local convolution, which is the core module in our model. We aim to formulate a kernel which is
\begin{itemize}
    \item location-characterized: In the local regions of different center nodes, the meteorological patterns governed by convolution kernel differ.
    \item smooth: Patterns are broadly similar when the center nodes are close in spatial distance.
    \item common: The kernel is shared by different local spaces where the neighbors' spatial distribution is distinct.
\end{itemize}
\paragraph{Kernel conditional on centers.} 
Contrary to Example 1 in DCRNN whose convolution kernel is predefined, we aim to propose the convolution kernel which can adaptively learn and imitate the location-characterized patterns of each local region centered at node $i$. A trivial way is to use a multi-layer neural network whose input is $\bm{\mathrm{x}}^{i'}$ to approximate the convolution kernel $\chi({\bm{\mathrm{x}}^{i'}})$. However, ${\bm{\mathrm{x}}^{i'}}$ as the input only represents the relative position and disables the kernel to capture the location-characterized patterns. For example, given two different center nodes whose neighbors' relative positions are totally the same, the convolution kernel in different locations will also coincide exactly, contrary to location-characterized patterns. 
Therefore, we propose to use conditional kernel, which reads $\chi({\bm{\mathrm{x}}^{i'}} ; \bm{\mathrm{x}}_{i})$, meaning that the convolution kernel in a certain local region is determined by the center node $\bm{\mathrm{x}}_{i}$. An multi-layer feedforward network is used to approximate this term, as
\begin{align}
    \chi({\bm{\mathrm{x}}^{i'}} ; \bm{\mathrm{x}}_{i}) = \mathrm{MLP}([\bm{\mathrm{x}}^{i'} , \bm{\mathrm{x}}_{i}]). \label{eq:mlpkernel}
\end{align}
\paragraph{Smoothness of local patterns.} We assume that the localized patterns of meteorological message flows have the property of \textbf{smoothness} -- two close center nodes' patterns of aggregation of messages from neighbors should be similar. In the light of convolution kernel, we define the smoothness of kernel function as follows:
\begin{Definition} 
The conditional kernel $\chi(\cdot|\cdot) $ is smooth, if it satisfies that for any $\epsilon > 0$, there exist $\delta>0$, such that for any two points $\bm{\mathrm{x}}_i, \bm{\mathrm{x}}_j \in {S^2}$ with $d_{S^2}(\bm{\mathrm{x}}_i, \bm{\mathrm{x}}_j) \leq \delta$,
\begin{align*}
\sup_{\bm{\mathrm{v}}\in \mathcal{C}_{\bm{\mathrm{x}}_i} {S^2},\bm{\mathrm{u}}\in \mathcal{C}_{\bm{\mathrm{x}}_j} {S^2}\atop \Pi_{\bm{\mathrm{x}}_i}(\bm{\mathrm{v}}) = \Pi_{\bm{\mathrm{x}}_j}(\bm{\mathrm{u}})}  |\chi(\Pi_{\bm{\mathrm{x}}_i}(\bm{\mathrm{v}});\bm{\mathrm{x}}_j) - \chi(\Pi_{\bm{\mathrm{x}}_j}(\bm{\mathrm{u}});\bm{\mathrm{x}}_i)|\leq \epsilon.
\end{align*}
\end{Definition}
The definition of smoothness of location-characterized kernel is motivated by the fact that if the distance between two center nodes $d_{S^2}(\bm{\mathrm{x}}_{i}, \bm{\mathrm{x}}_{j})$ is very small, the meteorological patterns in two local region should be of little difference, and thus kernel function $\chi(\cdot;\bm{\mathrm{x}}_{i})$ and $\chi(\cdot;\bm{\mathrm{x}}_{j})$ should be almost exactly the same.

The unified standard for choice of orthogonal basis on the cylindrical-tangent plane avoids problems caused by path-dependent parallel transport \cite{cohen2019gauge}, and contributes to the smoothness of conditional kernel. The property is likely to be compromised without unified  standard for orthogonal basis, as the following example illustrates.
\begin{Example}\rm  (shown in Fig.~\ref{fig:exampleunify}.)  For one node $\bm{\mathrm{x}}_{i}$, the orthogonal basis is $\{\mathbf{e}_{\phi_i}, \mathbf{e}_{z_i}\}$ as previously defined in Eq.~\ref{eq:orthobasis}, and for another node $\bm{\mathrm{x}}_{j}$ which is close to it, it is $\{-\mathbf{e}_{\phi_i}, \mathbf{e}_{z_i}\}$. There exists a node $\bm{\mathrm{x}}_{p}$ which is in their east on the sphere as the neighbors of both, and has great meteorological impacts on both of them. In one's local coordinate system, the first coordinate is positive while it is negative in another. Then if the kernel is smooth, the neighbor $\bm{\mathrm{x}}_{p}$ from the east will never be given large value in both local regions centered at node $i$ and $j$, violating the true patterns in meteorology, or it is likely to violate the smoothness of the kernel.
\end{Example}
As such, by using $\mathrm{MLP}(\cdot)$ as the approximator with smooth activate function like $\tanh$ and unifying the standard for choice of orthogonal basis, the smoothness property of conditional kernel can be ensured. However, the irregular spatial distribution of discrete nodes conflicts with the continuous kernel function shared by different center nodes, which will be discussed in the next part. 

\paragraph{Reweighting for irregular spatial distribution.}  Because the kernel function is continuous and shared by different center nodes, when the spatial distribution of each node's neighbors is similar or even identical, e.g. nodes are distributed on regular spatial grids in local spaces, the proposed conditional kernel takes both distance and orientation into consideration. However, the nodes are discrete and irregularly distributed on the sphere. Since the kernel is shared by all center nodes, the distinct spatial distribution of neighbors of different center nodes is likely to disrupt the smoothness of local patterns. An explicit example is given to illustrate the problems brought about by it.
\begin{Example}
\rm (shown in Fig.~\ref{fig:examplereweight}.) The two center nodes are close in distance, but the spatial distribution of their neighbors is different. The number of the right center's neighbors located in the south-west is two, while it is one for the left center. If the kernel is smooth, the message from the south-west flowing into the right center will be about twice than it from the south-west flowing into the left. 
\end{Example} 
To reweight the convolution kernel for each $\bm{\mathrm{x}}_{j} \in \mathcal{V}(i)$, we consider both their angle and distance scales. We first turn its representation in Cartesian coordinate system $\bm{\mathrm{x}}_j^{i'} = (\phi^{i'}_j, z^{i'}_j)$ in cylindrical-tangent space of $\bm{\mathrm{x}}_i$ into polar coordinate $(\varphi^{i'}_j, \rho^{i'}_j)$, where $\varphi^{i'}_j = \arctan (z^{i'}_j/\phi^{i'}_j)$ and  $\rho^{i'}_j = \sqrt{(z^{i'}_j)^2 + (\phi^{i'}_j)^2}$. Note that $\rho^{i'}_j$ equals to the geodesics induced distance between the two nodes on sphere.  In terms of angle, we calculate the \textbf{angle bisector} of every pair of adjacent nodes in the neighborhood according to $\varphi^{i'}_j$. We denote the angle between two adjacent angular bisectors of $\bm{\mathrm{x}}_j^{i'}$ by $\psi^{i'}_j$ (as shown in lower-right subfigure in Fig.~\ref{fig:examplereweight} ), and thus the angle scale is written as $\psi^{i'}_j/2\pi$. The distance scale is obtained similarly as DCRNN in Example 1, which reads $\exp(- (\rho^{i'}_j)^2/\tau)$, where $\tau$ is a learnable parameter.

To sum up, combining the two scaling terms with Eq.~\ref{eq:mlpkernel}, the final formulation of the smooth conditional local kernel in the case of irregular spatial distribution reads
\begin{align}
    \chi({\bm{\mathrm{x}}^{i'}_j} ; \bm{\mathrm{x}}_{i}) = \frac{\psi^{i'}_j}{2\pi}\exp(- \frac{(\rho^{i'}_j)^2}{\tau})\mathrm{MLP}([\bm{\mathrm{x}}^{i'}_j , \bm{\mathrm{x}}_{i}]). \label{eq:clckernel}
\end{align}

\subsection{4.4. Inapplicability in traffic forecasting}
\begin{figure}[ht]
\centering
\vspace{-0.3cm}
        \subfigure[Flows in meteorology]{ 
			\includegraphics[width=0.45\linewidth]{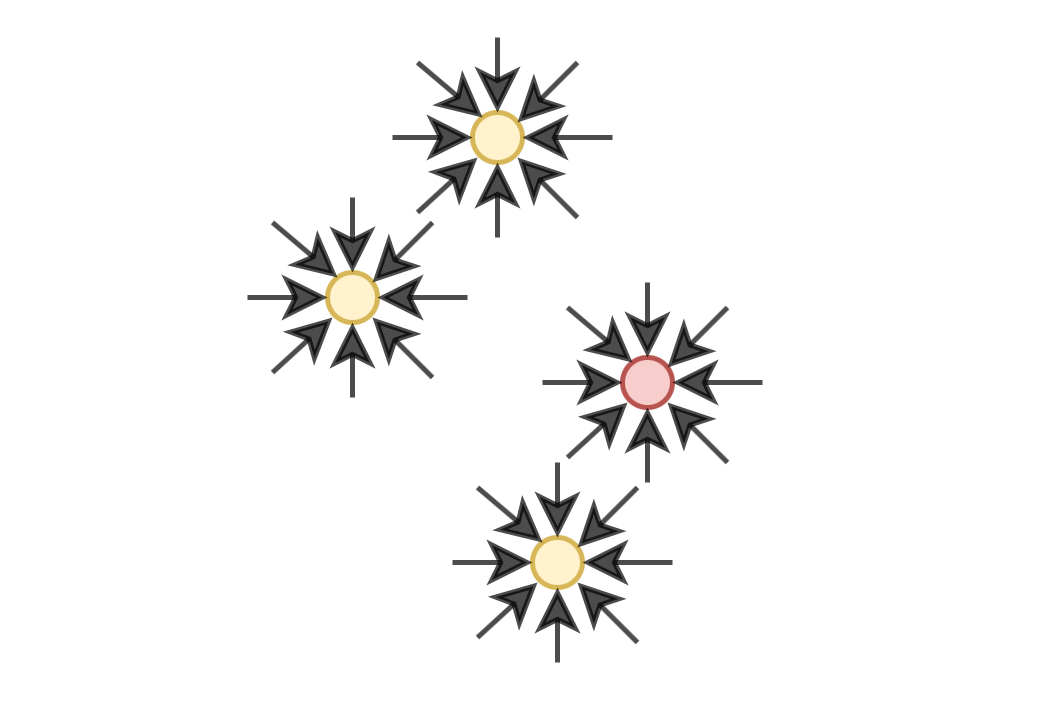}}\hspace{1mm}
		\subfigure[Flows in traffic]{
			\includegraphics[width=0.45\linewidth]{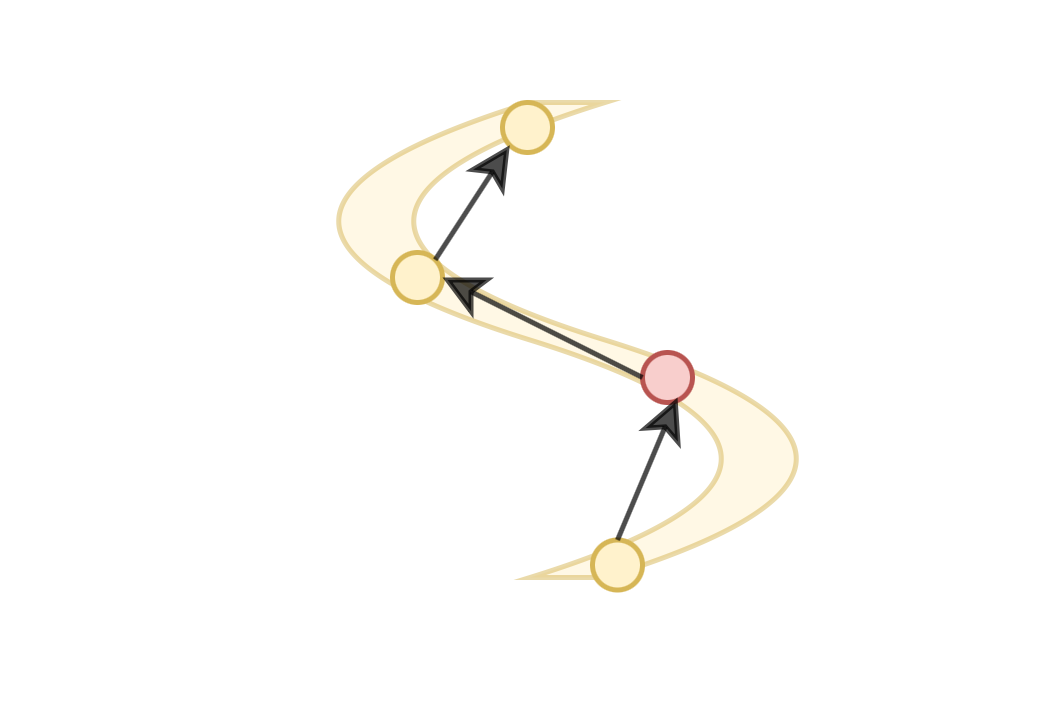}}\\\vspace{-0.3cm}
		\subfigure[Local pattern in meteorology]{
			\includegraphics[width=0.48\linewidth]{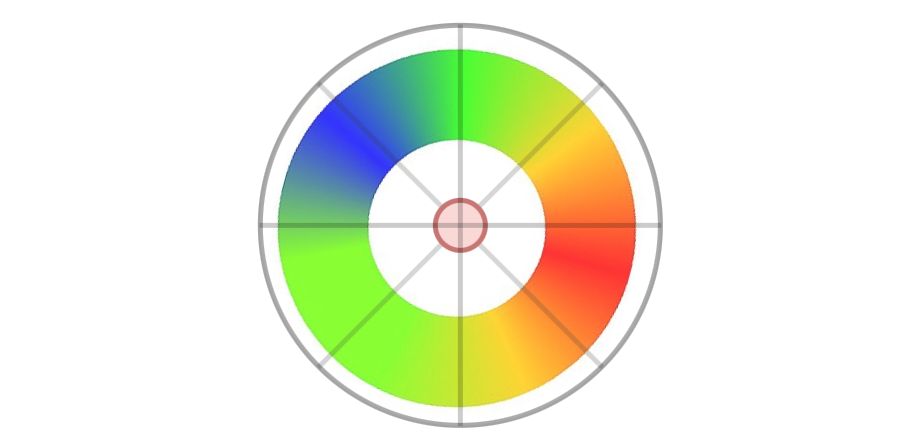}}\hspace{1mm}
		\subfigure[Local pattern in traffic]{
			\includegraphics[width=0.47\linewidth]{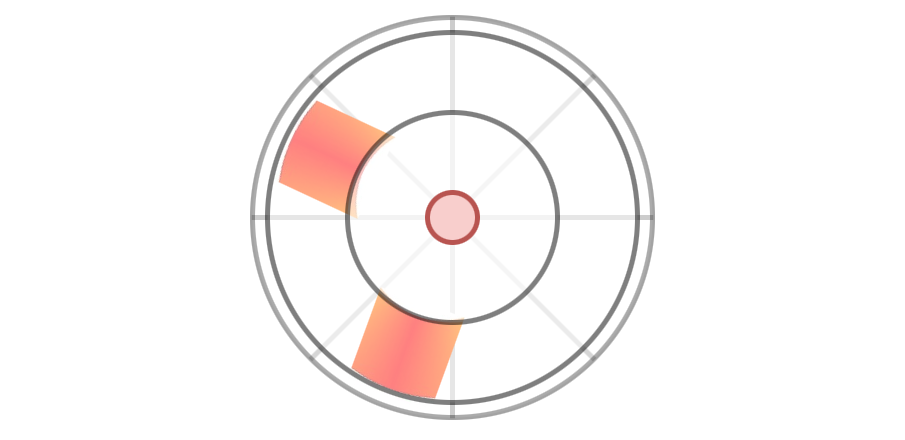}}\vspace{-0.3cm}
\caption{Different geographic sample spaces and local patterns in meteorology and traffic.}\vspace{-0.3cm}
\end{figure}

The proposed convolution is inapplicable to traffic forecasting. One reason is that the smoothness is not a reasonable property of local traffic flows' patterns, i.e. great difference may exist between traffic patterns of two close regions. An important transportation hub exit may exist in the middle of them, so the patterns are likely to differ a lot. Besides, because our convolution kernel is continuous in spatial domain, the continuity in orientation of the local convolution kernel is of no physical meaning in traffic irregular networks. In essence, the irregular structure of the road network restricts the flows of traffic to road direction, stopping vehicles from crossing the road boundary, so that the geographic sample space is restricted to the road networks, and traffic can only flow along roads. In comparison, the flows in meteorology like heat and wind can diffuse freely on the earth, without boundary, and the geographic sample space is the whole earth surface, enabling the local patterns to satisfy the continuity and smoothness.

\begin{figure*}[htb]
    \centering
    \includegraphics[width=6.2in]{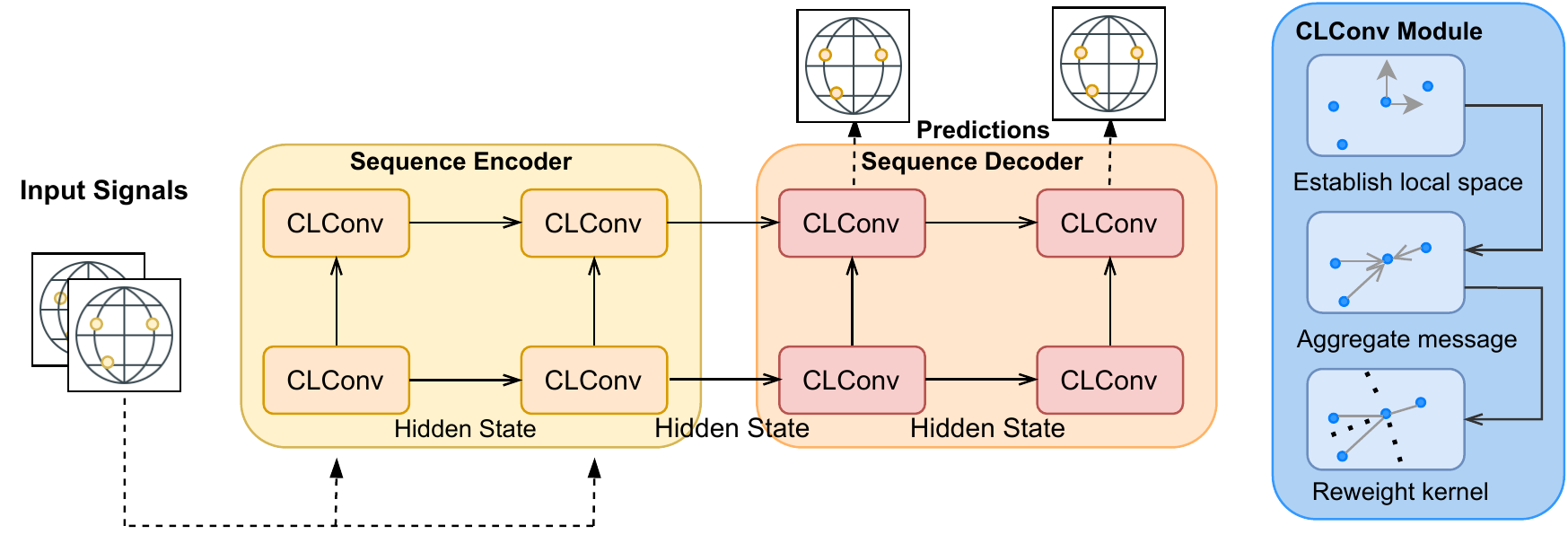}
    \vspace{-0.3cm} 
    \caption{Overall workflows and architecture of CLCRN.}
    \label{fig:overallarchi} 
\end{figure*}

\subsection{4.5. Temporal dynamics modeling}
The temporal dynamics is modeled as DCRNN does, which replaces fully-connected layers in cells of recurrent neural network with graph convolution layers. Using the kernel proposed in Eq.~\ref{eq:clckernel}, we obtain the GRU cell constituting the conditional local convolution recurrent network (CLCRN), whose overall architecture is shown in Fig.~\ref{fig:overallarchi}.

The overall neural network architecture for multi-step forecasting is implemented based on Sequence to Sequence framework \cite{Sutskever2014seq2seq}, with the encoder fed with previously observed time series and decoder generating the predictions. By setting a target function of predictions and ground truth observations such as minimal mean square error, we can use backpropagation through time to update the parameters in the training step. More details are given in Appendix C.

\section{5. Experiments}
\subsection{5.1. Experiment setting}
\begin{table}[htb]
    \resizebox{1\columnwidth}{!}{
    \begin{tabular}{cllcc}
        \toprule
    Methods & Spatial     & Temporal  & Learnable & Continuous  \\
    \midrule
    STGCN   & Vanilla GCN         & 1-D Conv           & \XSolidBrush               & \XSolidBrush             \\
    MSTGCN  & ChebConv              & 1-D Conv           & \XSolidBrush               & \XSolidBrush            \\
    ASTGCN  & GAT                     & Attention         & \CheckmarkBold             & \XSolidBrush           \\
    TGCN    & Vanilla GCN          & GRU               & \XSolidBrush               & \XSolidBrush               \\
    GCGRU   & ChebConv        & GRU               & \XSolidBrush               & \XSolidBrush                 \\
    DCRNN   & DiffConv            & GRU               & \XSolidBrush               & \XSolidBrush               \\
    AGCRN   & Node Similarity         & GRU               & \CheckmarkBold             & \XSolidBrush             \\
    CLCRN   & CondLocalConv     & GRU               & \CheckmarkBold          & \CheckmarkBold        \\
    \bottomrule    
    \end{tabular}
        }%
    \caption{Comparison of different spatio-temporal methods. \textit{`Spatial'} and \textit{`Temporal'}  represent the spatial convolution and temporal dynamics modules. If the spatial kernel is predefined, it is not \textit{`learnable'}. Only our method is established for \textit{`continuous'} spatial domain from which meteorological signals usually sampled.}\label{tab:modelcomparision}
\vspace{-0.4cm}\end{table}
\paragraph{Datasets.} The datasets used for performance evaluation are provided in \textbf{WeatherBench} \cite{rasp2020weatherbench}, with 2048 nodes on the earth sphere. We choose four hour-wise weather forecasting tasks including temperature, cloud cover, humidity and surface wind component, the units of which are $\mathrm{K}$, $\%\times 10^{-1}$, $\%\times 10$, $\mathrm{ms}^{-1}$ respectively. We truncate the temporal scale from Jan.1, 2010 to Dec.31, 2018, and set input time length as 12 and forecasting length as 12 for the four datasets. 
\paragraph{Metrics.} We compare CLCRN with other methods by deploying three widely used metrics - Mean Absolute Error (MAE), Root Mean Square Error (RMSE), and Mean Absolute Percentage Error (MAPE) to measure the performance of predictive models. 
\paragraph{Protocol.}Seven representative methods are set up, which can be classified into attention-based methods \cite{rozemberczki2021pytorch}: STGCN \cite{yu2018STGCN}, MSTGCN, ASTGCN \cite{Guo2019ASTGCN} and recurrent-based method: TGCN \cite{Zhao2020TGCN}, GCGRU \cite{seo2016gcgru}, DCRNN \cite{li2018diffusion}, AGCRN \cite{bai2020adaptive}. 
Note that the spatial dependency in AGCRN is based on the product of learnable nodes' embeddings, which is called 'Node Similarity'.
The comparison of these methods and ours are given in Table.~\ref{tab:modelcomparision}.
All the models are trained with target function of MAE and optimized by Adam optimizer for a maximum of 100 epoches. The hyper-parameters are chosen through a carefully tuning on the validation set (See Appendix D1 for more details). The reported results of mean and standard deviation are obtained through five experiments under different random seeds. 

\subsection{5.2 Performance comparison}
\begin{table*}[htb]
    \centering
    \resizebox{2.15\columnwidth}{!}{
    \begin{tabular}{c|ccccccccc|r}
    \toprule
    Datasets                     & Metrics & TGCN          & STGCN         & MSTGCN        & ASTGCN        & GCGRU      & DCRNN         & AGCRN         & CLCRN         & Improvements \\
    \midrule
    \multirow{2}{*}{Temperature} & MAE     & 3.8638±0.0970 & 4.3525±1.0442 & 1.2199±0.0058 & 1.4896±0.0130 & 1.3256±0.1499 & 1.3232±0.0864 & \underline{1.2551±0.0080} & \textbf{1.1688±0.0457} & 7.2001\%     \\
                                 & RMSE    & 5.8554±0.1432 & 6.8600±1.1233 & 1.9203±0.0093 & 2.4622±0.0023 & 2.1721±0.1945 & 2.1874±0.1227 & \underline{1.9314±0.0219} & \textbf{1.8825±0.1509} & 2.5318\%     \\
                                 \midrule
    \multirow{2}{*}{Cloud cover} & MAE     & 2.3934±0.0216 & 2.0197±0.0392 & 1.8732±0.0010 & 1.9936±0.0002 & \underline{1.5925±0.0023} & 1.5938±0.0021 & 1.7501±0.1467 & \textbf{1.4906±0.0037} & 6.3987\%     \\
                                 & RMSE    & 3.6512±0.0223 & 2.9542±0.0542 & 2.8629±0.0073 & 2.9576±0.0007 & 2.5576±0.0116 & \underline{2.5412±0.0044} & 2.7585±0.1694 & \textbf{2.4559±0.0027} & 3.3567\%     \\
                                 \midrule
    \multirow{2}{*}{Humidity}    & MAE     & 1.4700±0.0295 & 0.7975±0.2378 & 0.6093±0.0012 & 0.7288±0.0229 & \underline{0.5007±0.0002} & 0.5046±0.0011 & 0.5759±0.1632 & \textbf{0.4531±0.0065} & 9.5067\%     \\
                                 & RMSE    & 2.1066±0.0551 & 1.1109±0.2913 & 0.8684±0.0019 & 1.0471±0.0402 & \underline{0.7891±0.0006} & 0.7956±0.0033 & 0.8549±0.2025 & \textbf{0.7078±0.0146} & 10.3028\%    \\
                                 \midrule
    \multirow{2}{*}{Wind}        & MAE     & 4.1747±0.0324 & 3.6477±0.0000 & 1.9440±0.0150 & 2.0889±0.0006 & \underline{1.4116±0.0057} & 1.4321±0.0019 & 2.4194±0.1149 & \textbf{1.3260±0.0483} & 6.0640\%     \\
                                 & RMSE    & 5.6730±0.0412 & 4.8146±0.0003 & 2.9111±0.0292 & 3.1356±0.0012 & \underline{2.2931±0.0047} & 2.3364±0.0055 & 3.4171±0.1127 & \textbf{2.1292±0.0733} & 7.1475\%    \\
                                 \bottomrule
    \end{tabular}
    }
    \caption{MAE and RMSE comparison in forecasting length of $12\mathrm{h}$. Results with
    \underline{underlines} are the best performance achieved by baselines, and results with \textbf{bold} are the overall best. Comparisons in other lengths and metrics are shown Appendix D2. }\label{tab:comparison}\vspace{-0.3cm}
    \end{table*}
 Because MAPE is of great difference among methods and hard to agree on an order of magnitude, we show it in Appendix D2. 
\begin{figure}[ht]
\centering

        \subfigure[MAE on Temperature]{ 
			\includegraphics[width=0.47\linewidth]{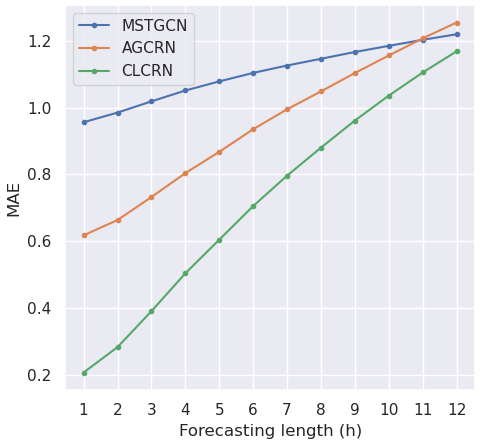}}\hspace{1mm}
		\subfigure[MAE on Cloud cover]{
			\includegraphics[width=0.47\linewidth]{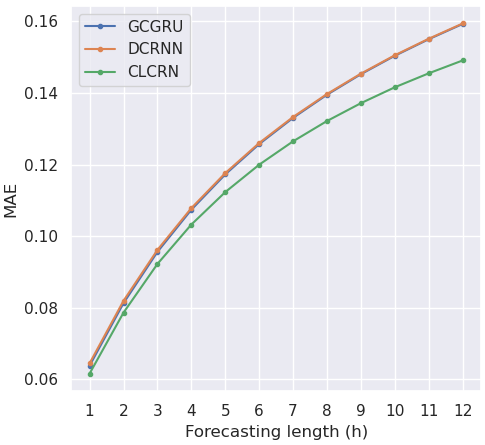}}\\\vspace{-0.2cm}
            \addtocounter{figure}{-1}
        \end{figure}
\begin{figure}
        \subfigure[MAE on Humidity]{
			\includegraphics[width=0.47\linewidth]{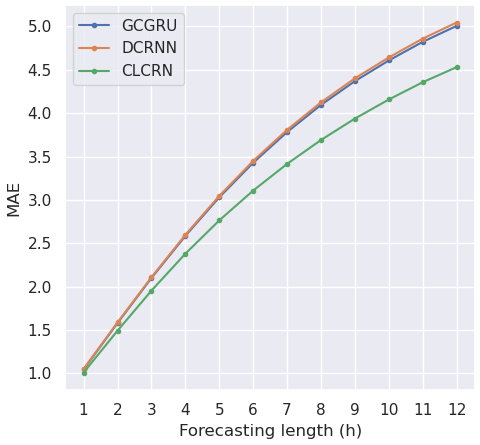}}\hspace{1mm}
		\subfigure[MAE on Wind]{
			\includegraphics[width=0.47\linewidth]{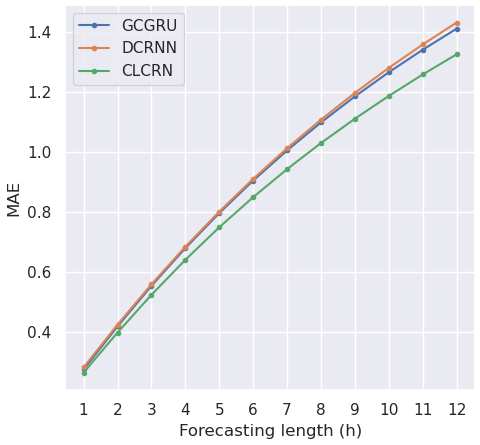}}\vspace{-0.2cm}
\caption{To avoid complicated and verbose plots, we choose the top three methods for MAE comparison in different forecasting length. }\label{fig:comparison}\vspace{-0.3cm}
\end{figure}

From Table.~\ref{tab:comparison} and Fig.~\ref{fig:comparison}, it can be conclude that 
(1) The recurrent-based methods outperform the attention-based, except that in Temperature dataset, MSTGCN works well.
(2) Our method further improves recurrent-based methods in weather prediction with a significant margin.
(3) Because most of the compared methods are established for traffic forecasting, they demonstrate a significant decrease in performance for meteorological tasks, such as TGCN and STGCN. The differences of the two tasks are analyzed by Sec. 4.4.
The `performance convergence' phenomenon on Temperature is explained in Appendix D2.

\subsection{5.3. Visualization of local patterns}
\begin{figure}[H]
    \centering
            \includegraphics[width=2.7in]{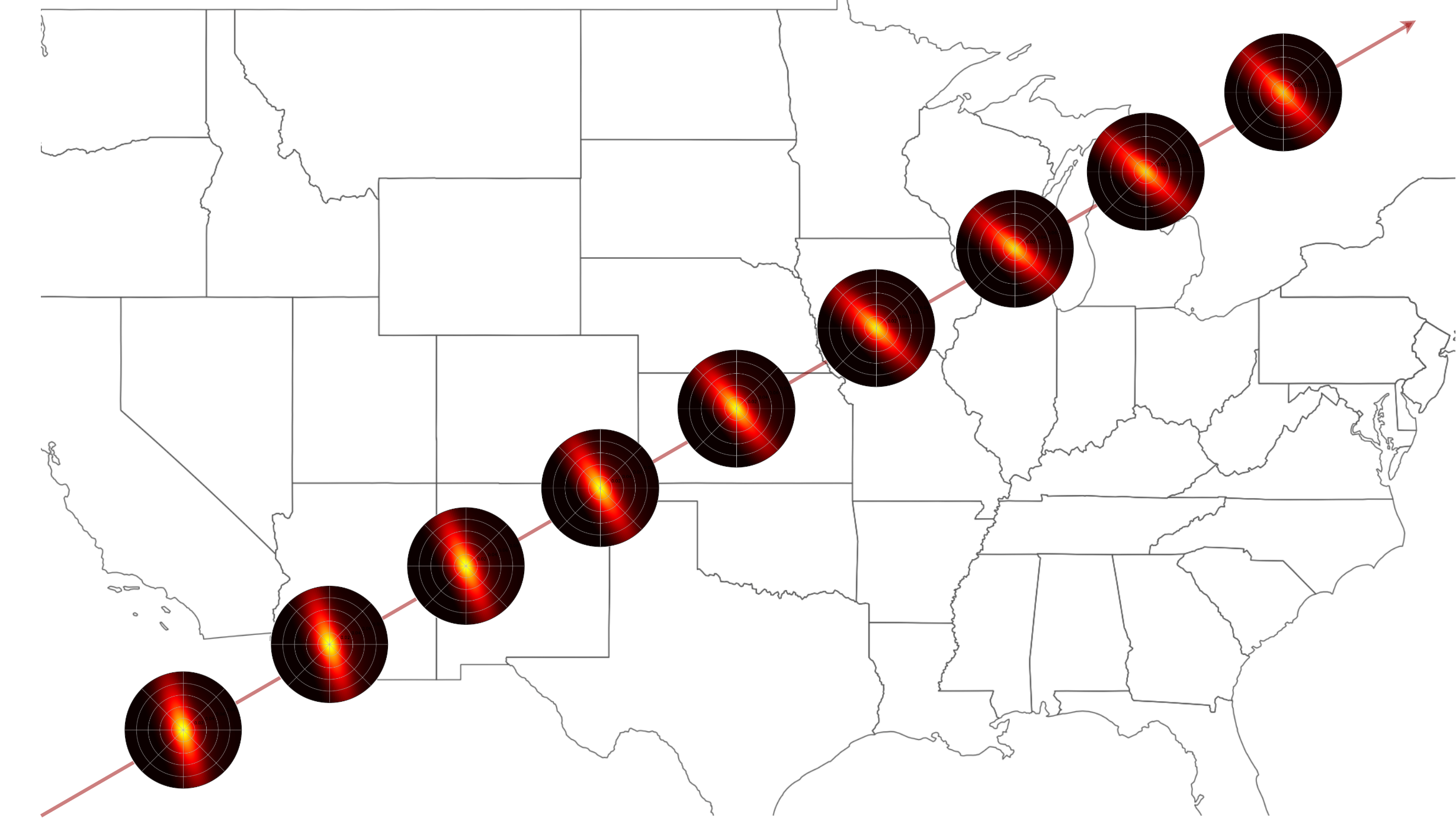}
    \caption{Changes of local kernels $\chi(\bm{\mathrm{x}}^{i'};\bm{\mathrm{x}}_{i})$ for uniformly-spaced $\bm{\mathrm{x}}_{i}$ obtained by trained CLCRN according to Humidity dataset.}\label{fig:patternchange}
    \end{figure}
The proposed convolution kernel aims to imitate the meteorological local patterns. For this, we give visualization of conditional local kernels to further explore the local patterns obtained from trained models. We choose a line from the south-west to the north-east in USA, and sample points as center nodes uniformly on the line.

As shown in Fig.~\ref{fig:patternchange}, the kernels conditional on center nodes show the smoothness property, and the patterns obtained from Humidity datasets demonstrate obvious directionality - nodes from the north-west and south-east impact the centers most. However, the kernel is over-smoothing - The change is very little although the center nodes vary a lot, which will be one of our future research issues.

\subsection{5.4. Framework choice: CNN or RNN?}
As concluded in (1) in performance comparison, recurrent-based methods usually outperform the attention-based in our evaluation. For the latter one, classical CNNs are usually used for intra-sequence temporal modeling. Here we further establish the CLCSTN by embedding our convolution layer into the framework of MSTGCN, as the attention-based version of CLCRN, to compare the cons and pros of the two frameworks. 
\begin{table}[htb]
\resizebox{1.05\columnwidth}{!}{
    \begin{tabular}{c|c|ccccc}
    \toprule
                    & Lengths              & Metrics & Temperature   & Cloud cover   & Humidity      & Wind          \\
                    \midrule
    \multirow{6}{*}{\rotatebox{90}{CLCSTN}} & \multirow{2}{*}{3h}  & MAE     & 1.1622±0.2773 & 1.5673±0.0050 & 0.4710±0.0423 & 1.2262±0.0072 \\
                            &                      & RMSE    & 1.9097±0.5892 & 2.4798±0.0105 & 0.6765±0.0596 & 1.8085±0.0163 \\ \cline{2-7} 
                            & \multirow{2}{*}{6h}  & MAE     & 1.2516±0.2762 & 1.6461±0.0052 & 0.5125±0.0401 & 1.3582±0.0070 \\
                            &                      & RMSE    & 2.0216±0.5409 & 2.5814±0.0106 & 0.7330±0.0553 & 1.9985±0.0168 \\ \cline{2-7} 
                            & \multirow{2}{*}{12h} & MAE     & 1.3325±0.2204 & 1.7483±0.0044 & 0.5691±0.0385 & 1.5727±0.0035 \\
                            &                      & RMSE    & 2.1239±0.3949 & 2.7101±0.0090 & 0.8104±0.0519 & 2.3058±0.0102 \\
                            \midrule
    \multirow{6}{*}{\rotatebox{90}{CLCRN}} & \multirow{2}{*}{3h}  & MAE     & 0.3902±0.0345 & 0.9225±0.0011 & 0.1953±0.0015 & 0.5233±0.0177 \\
                            &                      & RMSE    & 0.6840±0.0488 & 1.6428±0.0020 & 0.3307±0.0037 & 0.9055±0.0246 \\ \cline{2-7} 
                            & \multirow{2}{*}{6h}  & MAE     & 0.7050±0.0402 & 1.1996±0.0023 & 0.3107±0.0035 & 0.8492±0.0265 \\
                            &                      & RMSE    & 1.2408±0.1098 & 2.0611±0.0048 & 0.5114±0.0088 & 1.4296±0.0411 \\ \cline{2-7} 
                            & \multirow{2}{*}{12h} & MAE     & 1.1688±0.0457 & 1.4906±0.0037 & 0.4531±0.0065 & 1.3260±0.0483 \\
                            &                      & RMSE    & 1.8825±0.1509 & 2.4559±0.0027 & 0.7078±0.0146 & 2.1292±0.0733\\
    \bottomrule
    \end{tabular}\vspace{-0.3cm}
    }\caption{MAE and RMSE comparison in different forecasting length of CLCSTN and CLCRN.}\label{tab:clcstncom}
    \end{table}

\begin{figure}[htb] 
\centering
        \includegraphics[width=1.0\linewidth]{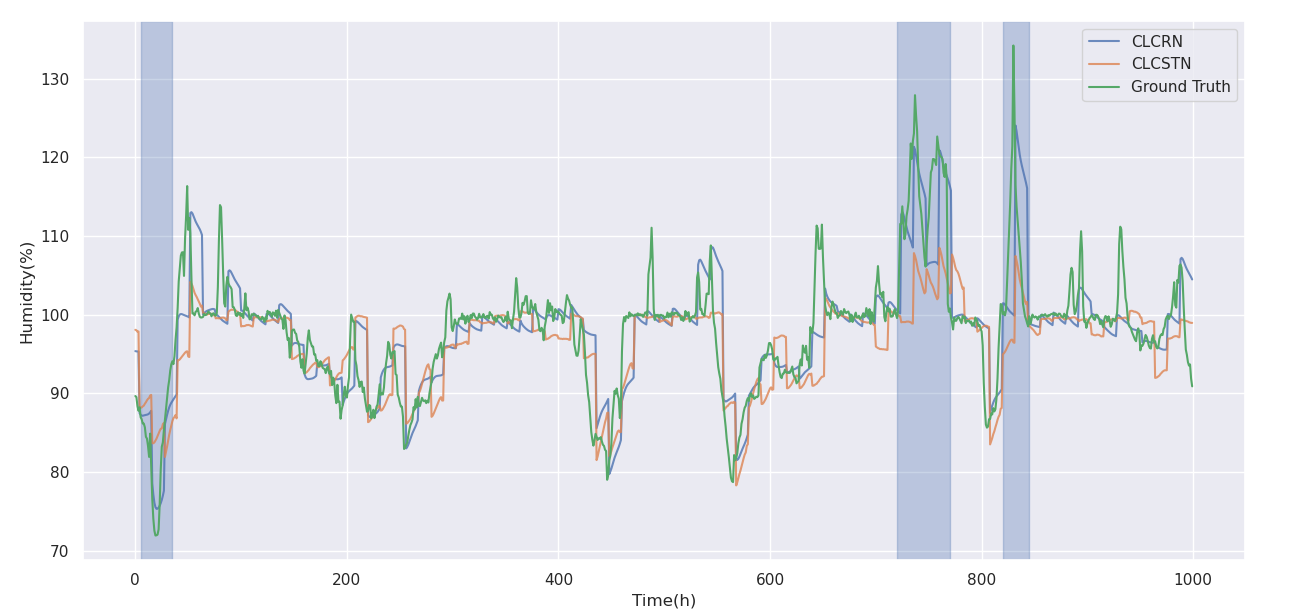}\vspace{-0.2cm}
        \caption{Predictions on Humidity, where the filled intervals show steep slopes and drastic fluctuations. (Appendix D3 in detail) }\label{fig:humidity}
\vspace{-0.3cm}
\end{figure}

\begin{figure}[H]
\includegraphics[width=1\linewidth]{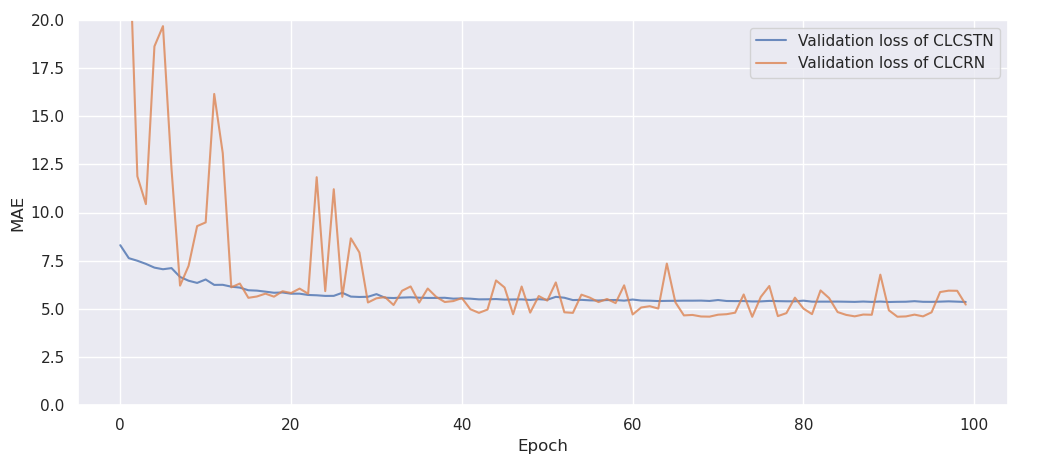}\vspace{-0.3cm}
\caption{Learning curve of the two methods on Humidity.} \label{fig:learningcurve}
\end{figure}

    From Table.~\ref{tab:clcstncom}, it is shown that the CLCRN outperforms CLCSTN in all evaluations. Besides, it is noted that the significant gap between two methods is in short term prediction rather than long term. We conjecture that the attention-based framework gives smoother prediction, while the other one can fit extremely non-stationary time series with great oscillation. Empirical studies given in Fig.~\ref{fig:humidity} show that the former framework tends to fit low-frequency signals, but struggles to fit short-term fluctuations. In long term, the influence of fitting deviation is weakened, so the performance gap is reduced. In this case, the fact that the learning curve of the former one is much smoother (Fig.~\ref{fig:learningcurve}) can be explained as well. The unstable learning curve is actually a common problem of all the recurrent-based models, which is another future research issue of ours.
\vspace{-0.7cm}
\subsection{5.5. Advantages of horizon maps}
\begin{table}[htb]\vspace{-0.3cm}
    \resizebox{1.05\columnwidth}{!}{
    \begin{tabular}{l|ccccc}
    \toprule
    Methods                                                                    & Metrics & Temperature    & Cloud cover   & Humidity      & Wind          \\
    \midrule
    CLCRN\_log & MAE     & 1.2638±0.1554 & 1.5599±0.0019 & 0.4663±0.0082 & 1.3958±0.0120 \\
               & RMSE    & 2.0848±0.1719 & 2.5171±0.0255 & 0.7341±0.0151 & 2.2659±0.0211 \\
                                                                                   \midrule
    CLCRN\_hor & MAE    & 1.1688±0.0457  & 1.4906±0.0037 & 0.4531±0.0065 & 1.3260±0.0483 \\
                                                                                   & RMSE   & 1.8825±0.1509  & 2.4559±0.0027 & 0.7078±0.0146 & 2.1292±0.0733\\
                                                                                   \bottomrule
    \end{tabular}
    
    }
    \caption{MAE and RMSE comparison in forecasting length of 12h for logarithmic and horizon maps.}\label{tab:mappingcomparison} 
In this part, we discussed two maps: logarithmic and horizon maps, which are established for two local spaces: tangent and cylindrical-tangent space, respectively. Here we compare the performance of our model with two different maps and spaces, to illustrate the advantages of the horizon maps as shown in Table.\ref{tab:mappingcomparison}.
\end{table}
\vspace{-0.3cm}
\subsection{5.6. Ablation study}
\vspace{-0.2cm}
\paragraph{Decomposition of the kernel.}
\begin{table}[htb]
    \resizebox{1.05\columnwidth}{!}{
    \begin{tabular}{l|ccccc}
    \toprule
    Composition                                                                    & Metrics & Temperature    & Cloud cover   & Humidity      & Wind          \\
    \midrule
    \multirow{2}{*}{Angle}                                                         & MAE    & 3.1673±0.3422  & 1.7787±0.0258 & 0.6653±0.0361 & 3.3753±0.4199 \\
                                                                                   & RMSE   & 4.8939±0.6142  & 2.8745±0.0572 & 1.0054±0.0721 & 5.1317±0.3603 \\
                                                                                   \midrule
    \multirow{2}{*}{Distance}                                                      & MAE    & 16.5671±0.0002 & 2.7308±0.0002 & 1.3443±0.0001 & 4.0531±0.0000 \\
                                                                                   & RMSE   & 21.7427±0.0085 & 3.7995±0.0029 & 1.8692±0.0003 & 5.2275±0.0004 \\
                                                                                   \midrule
    \multirow{2}{*}{MLP}                                                           & MAE    & 1.8815±0.1934  & 1.9047±0.0023 & 0.6208±0.1074 & 2.8672±0.0840 \\
                                                                                   & RMSE   & 2.9691±0.2311  & 3.1022±0.0215 & 0.9482±0.1496 & 4.2902±0.0982 \\
                                                                                   \midrule
    \multirow{2}{*}{\begin{tabular}[c]{@{}l@{}}MLP +\\ Distance\end{tabular}}      & MAE    & 1.4505±0.2248  & 1.8743±0.0038 & 0.6289±0.0711 & 2.5454±0.3485 \\
                                                                                   & RMSE   & 2.1754±0.2092  & 3.0627±0.0320 & 0.9388±0.1302 & 3.8815±0.5141 \\
                                                                                   \midrule
    \multirow{2}{*}{\begin{tabular}[c]{@{}l@{}}MLP + \\ Angle\end{tabular}}        & MAE    & \textbf{1.1205±0.2031}  & \underline{1.4919±0.0019} & \underline{0.4538±0.0064} & \textbf{1.2991±0.0344} \\
                                                                                   & RMSE   & \textbf{1.7957±0.2101}  & \textbf{2.4398±0.0110} & \underline{0.7082±0.0134} & \textbf{2.0763±0.0625} \\
                                                                                   \midrule
    \multirow{2}{*}{\begin{tabular}[c]{@{}l@{}}Angle +\\ Distance\end{tabular}}    & MAE    & 1.4986±0.0872  & 1.6907±0.0219 & 0.5378±0.0363 & 1.8932±0.2488 \\
                                                                                   & RMSE   & 2.1755±0.1042  & 2.7215±0.0330 & 0.8007±0.0484 & 3.0245±0.4769 \\
                                                                                   \midrule
    \multirow{2}{*}{\begin{tabular}[c]{@{}l@{}}MLP+Angle\\ +Distance\end{tabular}} & MAE    & \underline{1.1688±0.0457}  & \textbf{1.4906±0.0037} & \textbf{0.4531±0.0065} & \underline{1.3260±0.0483} \\
                                                                                   & RMSE   & \underline{1.8825±0.1509}  & \underline{2.4559±0.0027} & \textbf{0.7078±0.0146} & \underline{2.1292±0.0733}\\
                                                                                   \bottomrule
    \end{tabular}
    }
    \caption{MAE and RMSE comparison in forecasting length of 12h in different combinations of kernels. Results with \underline{underlines} are the best performance achieved by baselines, and results with \textbf{bold} are the overall best.}\label{tab:compositionkernel} \vspace{-0.3cm}
    \end{table}
As our kernel includes three terms shown in Eq.~\ref{eq:clckernel}, i.e. MLP term, distance scaling term and angle scaling term, we decompose the kernel to further validate the proposed kernel empirically.

From Table.~\ref{tab:compositionkernel}, we conclude that the `Distance' scaling term is of the least importance in that the performance of `MLP + Angle' is almost the same as it obtained by `MLP + Angle + Distance', and the kernel only composed of `Distance' usually performs worst.

\paragraph{Further analysis.}
There are several hyper-parameters (neighbor number $K$, number of layers and hidden units) determining the performance of our methods. We conduct experiments to explore their impacts on the performance empirically. More results are shown in Appendix D4. 
\begin{figure}[H] \vspace{-0.4cm}

\subfigure[Impacts of $K$]{ 
			\includegraphics[width=0.485\linewidth]{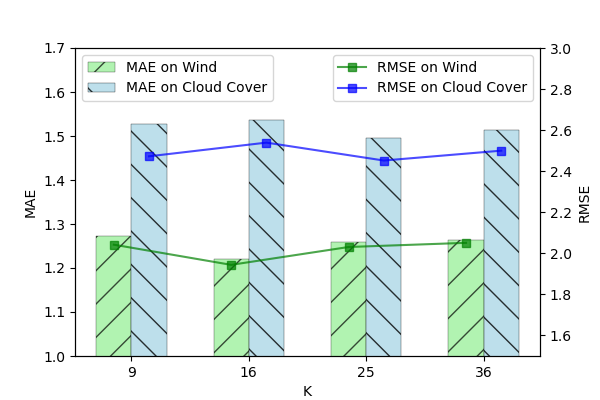}}
\subfigure[Impacts of layer number]{ 
			\includegraphics[width=0.485\linewidth]{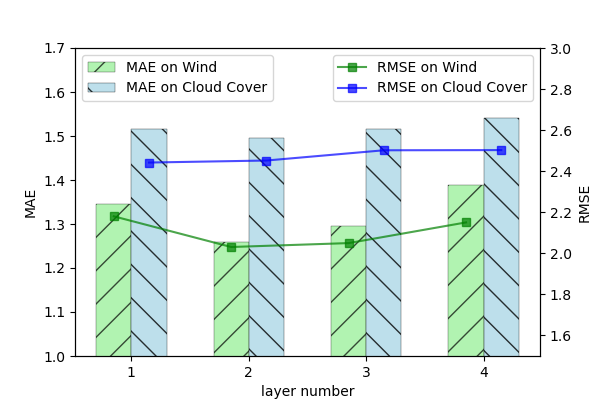}}
\vspace{-0.3cm}
\caption{Impacts of hyper-parameters $K$ and layer number on performance on two datasets. } \label{fig:hyperpara}
\end{figure}

\section{6. Conclusion}
We proposed a local conditional convolution to capture and imitate the meteorological flows of local patterns on the whole sphere, which is based on the assumption: smoothness of location-characterized patterns. 
An MLP and reweighting terms with continuous relative positions of neighbors and center as inputs are employed as convolution kernel to handle uneven distribution of nodes.

Empirical study shows the method achieves improved performance. Further analysis reveals two existing problems of our method: the \textbf{over-smoothness} of the learned local patterns (Sec. 5.3.) and \textbf{instability} of the training process (Sec. 5.4.), which would be our future research issues to focus on. 
\section{Acknowledgement}

CAIRI Internal Fund established by Prof. Stan. Z. Li provided support to assist the authors with research. Besides, Prof. Ling Li and Dr. Lu Yi shared expertise and insights in meteorology.

As a newcomer to the field of machine learning and artificial intelligence, I hope that I can make tiny but recognized contributions to noble scientific issues such as climate change. I will keep my enthusiasm with inspiration and diligence in my future academic life. Wish me good luck!
\clearpage
\bibliography{aaai22}
\clearpage
\appendix
\section{A: Notation and Formal Definition}
\subsection{A1: Glossary of notations}

\begin{table}[H]
    \centering
    \caption{Glossary of notations used in this paper.}
    \begin{tabular}{lp{0.65\linewidth}}
    \toprule
    Symbol                                                                                                                                  & Used for      \\    
    \midrule
    $t$                                                                                                                                     & Timestamp.    \\
    $N$                                                                                                                                     & Number of nodes sampled on sphere. \\
    $\mathcal{G} = (\mathcal{V}, \mathcal{E}, \bm{A})$                                                                                      & Graph represented nodes, edges and adjacency matrix respectively.\\                                                                            
    $\bm{\mathrm{x}}_i^S$                                                                                                                   & Coordinate representation of the $i$-th nodes on sphere.\\
    $\bm{\mathrm{x}}_i^E$                                                                                                                   & Coordinate representation of the $i$-th nodes on Euclidean space.\\
    $\bm{F}^{(t)}, \bm{F}^{(t)}_i$                                                                                                          & Signal Matrix at time $t$, and it of the $i$-th node.\\
    $\mathcal{N}(i)$                                                                                          & The set of neighbors of center node $i$.\\
    $\bm{h}^{l}_i$                                                                                                                          & The $i$-th nodes' representation after the $l$-th layer.\\
    $\mathcal{V}(i)$                                                                                                                        & The neighborhood coordinate set of center node $i$.\\
    $\star_{\mathcal{N}(i)}$                                                                                                                & Convolution on the $i$-th nodes' neighborhood.\\
    $\Omega(\bm{\mathrm{x}}^S_j, \bm{\mathrm{x}}^S_i)$                                                                                      & Convolution kernel measuring impacts of the $i$-th node on $j$.\\        
    $\chi(\bm{\mathrm{x}}^{i'})$                                                                                                            & Convolution kernel with relative position as input.\\        
    $\bm{H}(\bm{\mathrm{x}}^S_i)$                                                                                                           & Function mapping each point  on  sphere  to  its  feature  vector.\\        
    $\mathcal{L}_{\bm{\mathrm{x}}}S^M$                                                                                                      & Local space centering at $\bm{\mathrm{x}}$ on sphere.\\        
    $\mathcal{T}_{\bm{\mathrm{x}}}S^M$                                                                                                      & Tangent space centering at $\bm{\mathrm{x}}$ on sphere.\\        
    $\mathcal{C}_{\bm{\mathrm{x}}}S^M$                                                                                                      & Cylindrical-tangent space centering at $\bm{\mathrm{x}}$ on sphere.\\        
    $d_{S^M}(\bm{\mathrm{x}}, \bm{\mathrm{y}})$                                                                                             & Distance between $\bm{\mathrm{x}}$ and $\bm{\mathrm{y}}$ on sphere induced by geodesics.\\        
    $ \mathcal{M}_{\bm{\mathrm{x}}}(\cdot)$                                                                                                 & Isometric map on $\bm{\mathrm{x}}$'s local space.\\        
    $ \log_{\bm{\mathrm{x}}}(\cdot)$                                                                                                        & Logarithmic map on $\bm{\mathrm{x}}$'s tangent space.\\        
    $ \mathcal{H}_{\bm{\mathrm{x}}}(\cdot)$                                                                                                 & Horizon map on $\bm{\mathrm{x}}$'s cylindrical-tangent space.\\        
    $\Pi_{\bm{\mathrm{x}}}(\cdot)$                                                                                                          & Transform of point in local space to local coordinate system.\\        
    $P_{{\bm{\mathrm{x}}_i}}(\cdot)$                                                                                              & Normalized Projection operator.\\        
    $\bm{e}_{x}, \bm{e}_{y}, \ldots$                                                                                                        & Orthogonal basis.\\        
    $(\phi^{i'}, z^{i'} )$                                                                                                                  & Neighbors in local space represented by Cartesian coordinate system.\\        
    $(\varphi^{i'}, \rho^{i'})$                                                                                                             & Neighbors in local space represented by polar coordinate system.\\        
    $\psi^{i'}$                                                                                                                             & The angle between two adjacent angular bisectors in polar coordinate system.\\        
    \bottomrule
\end{tabular}
\end{table}
\subsection{A2: Formal Definition of Local Space}

\begin{Definition}
A manifold is a topological space that locally resembles Euclidean space near each point. More precisely, an n-dimensional manifold is a topological space with the property that each point has a neighborhood that is homeomorphic to an open subset of n-dimensional Euclidean space.\footnote{\url{https://en.wikipedia.org/wiki/Manifold}}
\end{Definition}

\begin{Proposition}
The M-dimension sphere is a manifold. Therefore, for each point $\bm{\mathrm{x}} \in S^M$, and a ball $\mathrm{B}_{S^M}(\bm{\mathrm{x}}, r) = \bm{\mathrm{y}}\in S^M:d_{S^M}(\bm{\mathrm{x}},\mathrm{\bm{y}}) < r\}$, there exist a homeomorphism $\phi$, such that $\phi: B_{S^M}(\bm{\mathrm{x}}, r) \rightarrow \mathbb{R}^M$ and $\max{r} = \pi$. In this way, we define the local space of $\bm{\mathrm{x}}$ as $\phi(\mathrm{B}_{S^M}(\bm{\mathrm{x}}, \pi)) = \mathcal{L}_{\bm{\mathrm{x}}}S^M$.
\end{Proposition}

\section{B: Local Space and Mappings}
\subsection{B1: Proof of Proposition 2}
\textit{Proof:}
It is easy to prove for any vector  $\bm{\mathrm{x}}$, after projection operator it satisfies $<P_{{\bm{\mathrm{x}}_i}}(\bm{\mathrm{x}}), \bm{\mathrm{x}}_i> = 0 $, where $P_{{\bm{\mathrm{x}}_i}}(\bm{\mathrm{x}}) = \frac{\bm{\mathrm{x}}}{||\bm{\mathrm{x}}||} - <\frac{\bm{\mathrm{x}}_i}{||\bm{\mathrm{x}}_i||},\frac{\bm{\mathrm{x}}}{||\bm{\mathrm{x}}||}>\frac{\bm{\mathrm{x}}_i}{||\bm{\mathrm{x}}_i||}$.

Then we first aim to prove that the vector after being mapped is located in cylindrical-tangent space of $\bm{\mathrm{x}_i}$. Denote $\bm{\mathrm{v}}_j = \mathcal{H}_{\bm{\mathrm{x}}_i}(\bm{\mathrm{x}}_j) $, then
\begin{align}
    &<\bm{\mathrm{v}}_j^-, \bm{\mathrm{x}}_i^->\\
    =& \frac{d_{S^2}(\bm{\mathrm{x}}_i, \bm{\mathrm{x}}_j)}{||[P_{{\bm{\mathrm{x}}_i^-}}(\bm{\mathrm{x}}_j^- - \bm{\mathrm{x}}_i^-), x_{j,3} - x_{i,3}]||}<P_{{\bm{\mathrm{x}}_i^-}}(\bm{\mathrm{x}}_j^- - \bm{\mathrm{x}}_i^-),\bm{\mathrm{x}}_i^->\\
    =& 0.
\end{align}
Finally, we prove that it is an isometric mapping, since 
\begin{align}
    &\left\lVert d_{S^2}(\bm{\mathrm{x}}_i, \bm{\mathrm{x}}_j)\frac{[P_{{\bm{\mathrm{x}}_i^-}}(\bm{\mathrm{x}}_j^- - \bm{\mathrm{x}}_i^-), x_{j,3} - x_{i,3}]}{||[P_{{\bm{\mathrm{x}}_i^-}}(\bm{\mathrm{x}}_j^- - \bm{\mathrm{x}}_i^-), x_{j,3} - x_{i,3}]||}\right\rVert\\
    =& d_{S^2}(\bm{\mathrm{x}}_i, \bm{\mathrm{x}}_j).
\end{align}
Therefore, the horizon map shown in Proposition 2. is an isometric map of nodes on $S^2$ into $\mathcal{C}_{\bm{\mathrm{x}}_i} S^2$.
\subsection{B2: Fast implementation and transforms on poles}
\paragraph{Fast implementation.} When the spatial sampling is dense on the sphere, we can also use a fast implementation to replace the local coordinate transform and horizon map, which reads 
\begin{equation}
    \bm{\mathrm{x}}^{i'}_j = 
    \begin{cases}
      &(\theta_j - \theta_i, \phi_j - \phi_i)   \quad \quad \quad  \phi_j - \phi_i \in [-\pi, \pi];\\
      &(\theta_j - \theta_i, \phi_j - \phi_i - 2\pi) \quad  \phi_j - \phi_i \in (\pi,2\pi);\\
      &(\theta_j - \theta_i, \phi_j - \phi_i + 2\pi) \quad  \phi_j - \phi_i \in (-2\pi,-\pi),\\
    \end{cases}
    \end{equation}
The empirical evaluation shows that it can also give a competitive performance, although the mapping just perserves the relative orientation on graticules, without considering the distance.
We infer that the reason for the phenomenon is that the inputs of the conditional local kernel is of the same scale. Further research will be conducted on this.

\paragraph{Handling the poles.} We use the negative relative positions of the pole in each neighbors local space as the neighbors' relative position. To be more specific, all the neighbors' first coordinate is $0$ in both poles, and the second are all negative in the North pole, while they are all positive in the South pole.

\section{C: Temporal Dynamics Modeling}
Temporal dynamics of node $i$ modeled by the GRU units with Conditional local convolution is shown as follows:
\begin{equation}
\label{eq:clcgru}
\begin{cases}
    & \bm{r}_i^{(t)} = \sigma(\Omega\star_ {{\mathcal{V}}(i)}[\bm{F}_i^{(t)}, \bm{Z}_i^{(t-1)}]\bm{W}_r + \bm{b}_r) ;\\
    & \bm{u}_i^{(t)} = \sigma(\Omega\star_ {{\mathcal{V}}(i)}[\bm{F}_i^{(t)}, \bm{Z}_i^{(t-1)}]\bm{W}_u + \bm{b}_u) ;\\
    & \bm{C}_i^{(t)} = \text{tanh}(\Omega\star_ {{\mathcal{V}}(i)}[\bm{F}_i^{(t)}, (\bm{r}_i^{(t)} \odot \bm{Z}_i^{(t-1)})]\bm{W}_C + \bm{b}_C);\\
    & \bm{Z}_i^{(t)} = \bm{u}_i^{(t)} \odot \bm{Z}_i^{(t-1)} + (\bm{I}-\bm{u}_i^{(t)}) \odot \bm{C}_i^{(t)},\\
\end{cases}
\end{equation}
where $\bm{W}_r, \bm{W}_u, \bm{W}_C$ are weights, and $\bm{b}_r, \bm{b}_u, \bm{b}_C$ are bias. $\odot$ is the Hadamard product, $\bm{F}_i^{(t)}$ is the input of the unit of node $i$ at time $t$, $\bm{Z}_i^{(t)}$ is the hidden state of node $i$ at time $t$ as well as the output of unit at time $t-1$, $\bm{r}^{(t)}$ and $\bm{u}^{(t)}$ is the reset and update gate defined in GRU respectively. In detail,
\begin{align*}
    \Omega\star_ {{\mathcal{V}}(i)}\bm{h}_i^{(t)} =   \sum_{\bm{\mathrm{x}}} \chi(\bm{\mathrm{x}}^{i'}) \bm{H}(\bm{\mathrm{x}})\delta_{\mathcal{V}(i)}(\bm{\mathrm{x}})
\end{align*}
is the convolution on $i$'s local space, where the kernel $\chi(\bm{\mathrm{x}}^{i'})$ is formulated by Eq.~\ref{eq:clckernel}. \section{D: Experiments}

\subsection{D1: Experiment protocol and details}
\textbf{Training details.}
All the methods are trained using the random seeds chosen in $\{2021, 2022, 2023, 2024, 2025\}$, from which the reported mean and standard deviation are obtained.
The basic learning rate is choosen as 0.01, and it decays with the ratio 0.05 per 10 epoch in the first 50 epoch.  Early-stopping techniques are used for choosing epoch according to validation set. The baselines are all implemented according to Geometric Temporal \cite{rozemberczki2021pytorch}.
Batch size for training is set as 32, and the ablation study in D4 only changes the target hyper-parameter. For example, when studying the effects of $K$, we set it as several different values, with others like layer numbers and hidden units unchanged.
The Pytorch framework is used, of the version 1.8.0 with CUDA version 11.1.  Every single experiment is executed on a server with one NVIDIA V100 GPU card, equipped with 32510 MB video memory.\\
\textbf{Hyper-parameters.}
In our method, the 3-layer MLP to approximate the convolution kernel is set up with the neuron number $[10,8,6]$, which is shared by all CLCGRU layers.\\
\textbf{Dataset details.} 
For more details for the four datasets in Weatherbench, the statistics is shown in Table.~\ref{tab:dataset}.\\
\begin{table}[htb]\vspace{-0.3cm}
    \resizebox{1.05\columnwidth}{!}{
        \begin{tabular}{c|cccc}
            \toprule
            Datasets           & Temperature & Cloud cover & Humidity & Wind              \\
            \midrule
            Dimension          & 1           & 1           & 1        & 2                 \\
            Input length       & 12          & 12          & 12       & 12                \\
            Forecasting length & 12          & 12          & 12       & 12                \\
            \#Nodes            & 2048        & 2048        & 2048     & 2048              \\
            \#Training set     & 2300        & 2300        & 2300     & 2300              \\
            \#Validation set   & 329         & 329         & 329      & 329               \\
            \#Test set         & 657         & 657         & 657      & 657               \\
            Mean value         & 278.8998    & 0.6750      & 78.0328  & -0.0921/0.2398    \\
            Max value          & 323.7109    & 1.0000      & 171.1652 & 32.2456/32.8977   \\
            Min value          & 193.0290    & 0.0000      & -2.4951  & -37.7845/-31.6869 \\
            Std value          & 21.1159     & 0.3617      & 18.2822  & 5.6053/4.8121    \\
            \bottomrule    
        \end{tabular}
    }\vspace{-0.3cm}
    \caption{Dataset statistics}\label{tab:dataset} \vspace{-0.3cm}
    \end{table}\vspace{-0.3cm}
\\
\paragraph{Metrics computation.}
Let $\mathbf{F}^{(t,T)} = [F^{(t)}, \ldots, F^{(t+T)}]$ be the ground truth, and $\mathbf{\hat F}^{(t,T)} = [\hat F^{(t)}, \ldots, \hat F^{(t+T)}]$ be the predictions given by neural networks. The three metrics including MAE, RMSE and MAPE are calculated as
\begin{align*}
    \mathrm{MAE}(\mathbf{F}^{(t,T)},\mathbf{\hat F}^{(t,T)}) &= \frac{1}{T}\sum_{i=t}^{T} |\hat F^{(i)} - F^{(i)} |\\
    \mathrm{RMSE}(\mathbf{F}^{(t,T)},\mathbf{\hat F}^{(t,T)}) &= \frac{1}{T}\sqrt{\sum_{i=t}^{T} |\hat F^{(i)} - F^{(i)} |^2}\\
    \mathrm{MAPE}(\mathbf{F}^{(t,T)},\mathbf{\hat F}^{(t,T)}) &= \frac{1}{T}\sum_{i=t}^{T} \frac{|\hat F^{(i)} - F^{(i)} |}{|F^{(i)} |}\\
\end{align*}
MAPE metrics is not stable, because there exists a term in the denominator, and thus we do not consider the contributions of terms with ground truth equalling 0. However, for Cloud cover, the minimal is still exrtremely small, causing the MAPE term extremely large.

\subsection{D2: Overall method comparison}
The model comparison of hour-wise prediction on MAE, RMSE and MAPE is shown in Fig. ~\ref{fig:overallcomp}.
For each dataset, we give the comparison of  methods with overall top-3 performance.

The reason for `convergence' of model performance on Temperature can be explained by visualization in Appendix D.3 and analysis in section 5.4. As we state in section 5.4, the drastic fluctuations are very hard for neural network to model. The plot of ground truth temperature in Figure. 8(a) is much smoother than other three datasets (Figure. 8), without drastic fluctuations. When the true values always fluctuate slightly around, the model giving smooth outputs as the future prediction usually performs well because the mean predictive errors are minor when the time scale is large enough. However, when the model tries to capture the fluctuations, even a single significant deviation from the ground truth causes the error to explode. In this way, the performances of models seem to ‘converge’ when time length increases on Temperature dataset, because the increase in time length benefits models with smooth outputs and hurts models trying to fit the fluctuations. 
\subsection{D3: More visualization of prediction}
We give more visualization to demonstrate the claim that attention-based methods is hard to fit the sharp fluctuations, as shown in Fig. ~\ref{fig:timeseriespred}.
\begin{figure}[H]
    \centering
    \subfigure[Predictions on Temperature on $1000$-th node.]{ 
        \includegraphics[width=1.0\linewidth]{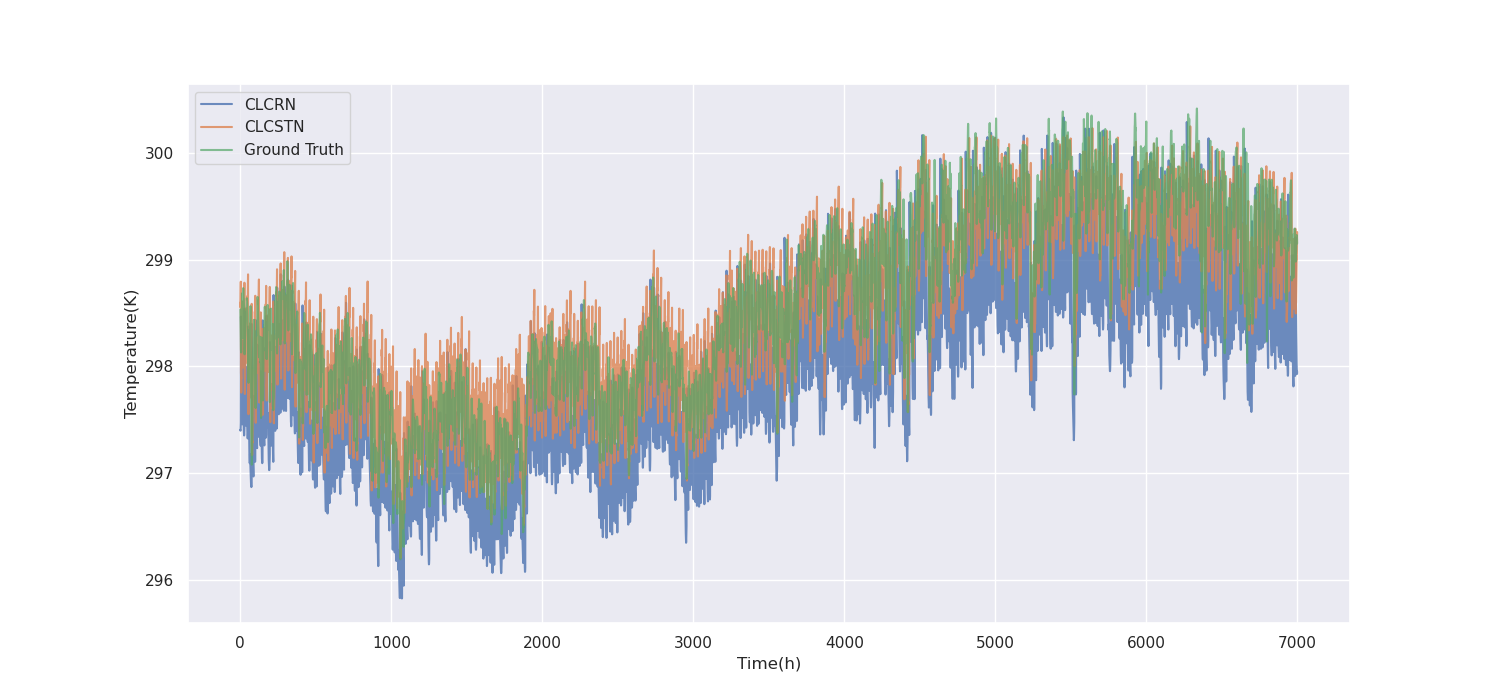}}\\
    \subfigure[Predictions on Cloud cover on $1000$-th node.]{ 
        \includegraphics[width=1.0\linewidth]{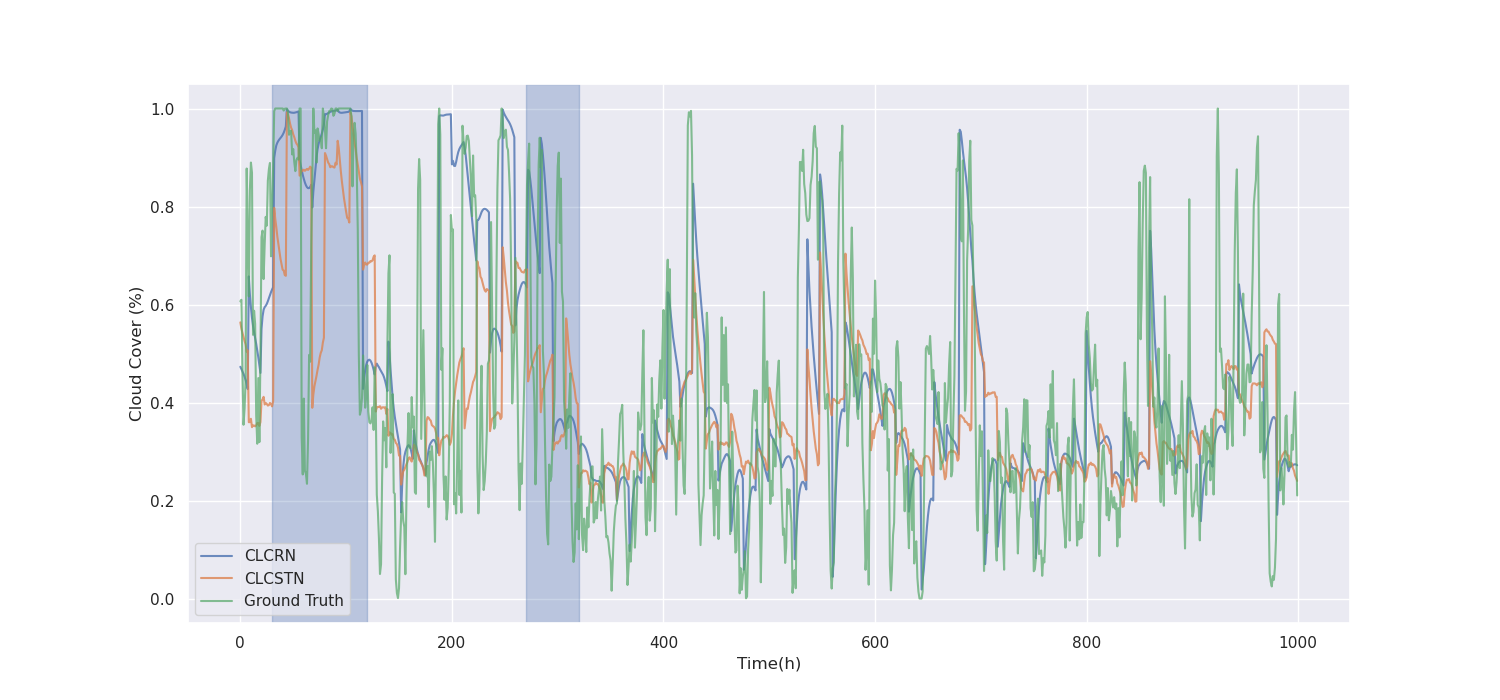}}\\
    \subfigure[Predictions on Cloud cover on $2000$-th node.]{ 
        \includegraphics[width=1.0\linewidth]{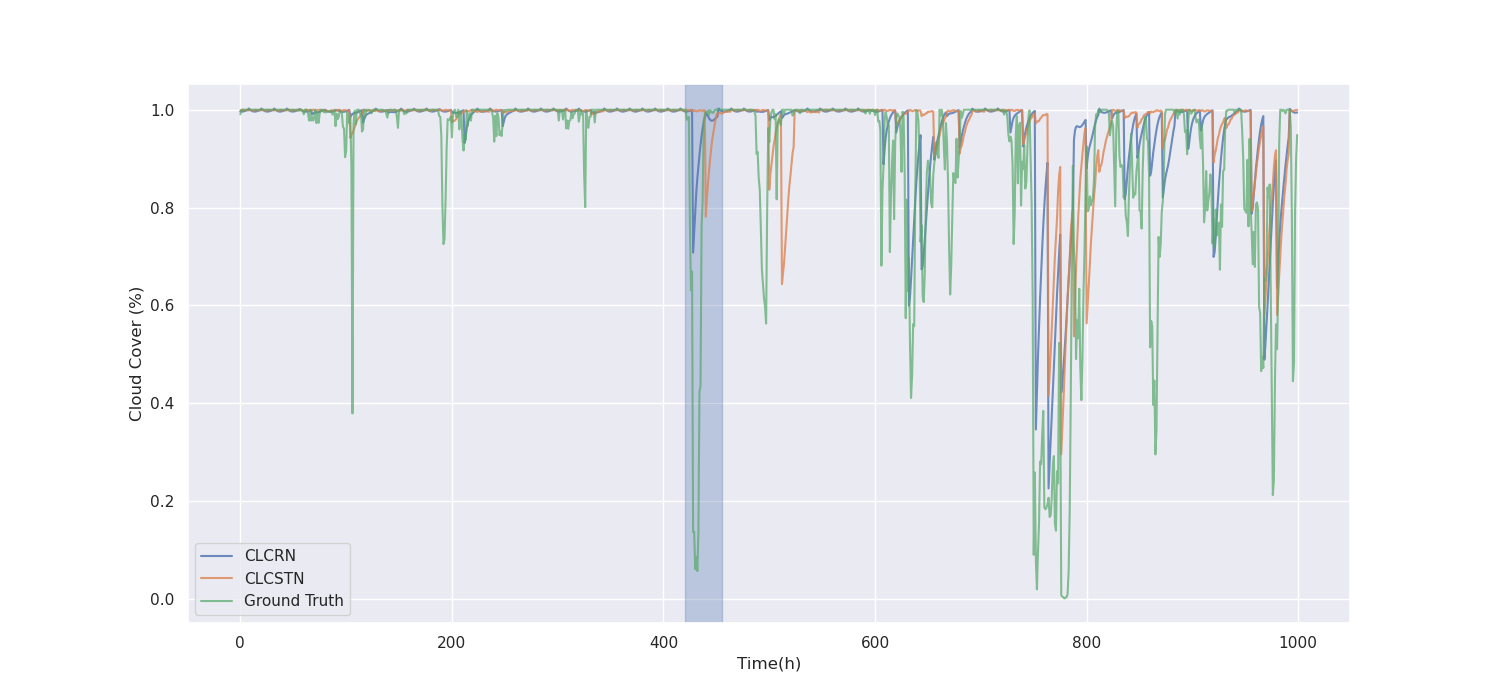}}\\
    \subfigure[Predictions on Humidity on $1000$-th node.]{ 
        \includegraphics[width=1.0\linewidth]{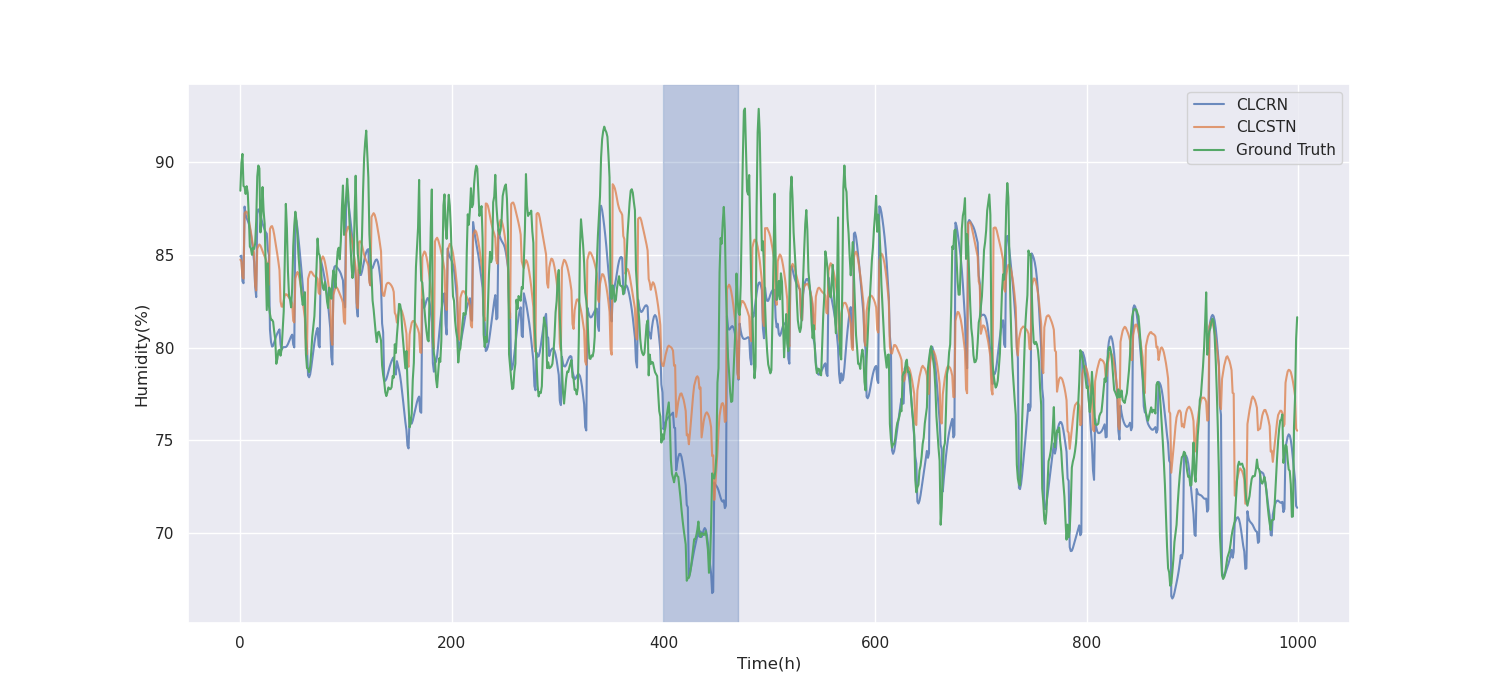}}\\
    \subfigure[Predictions on Humidity on $2000$-th node.]{ 
        \includegraphics[width=1.0\linewidth]{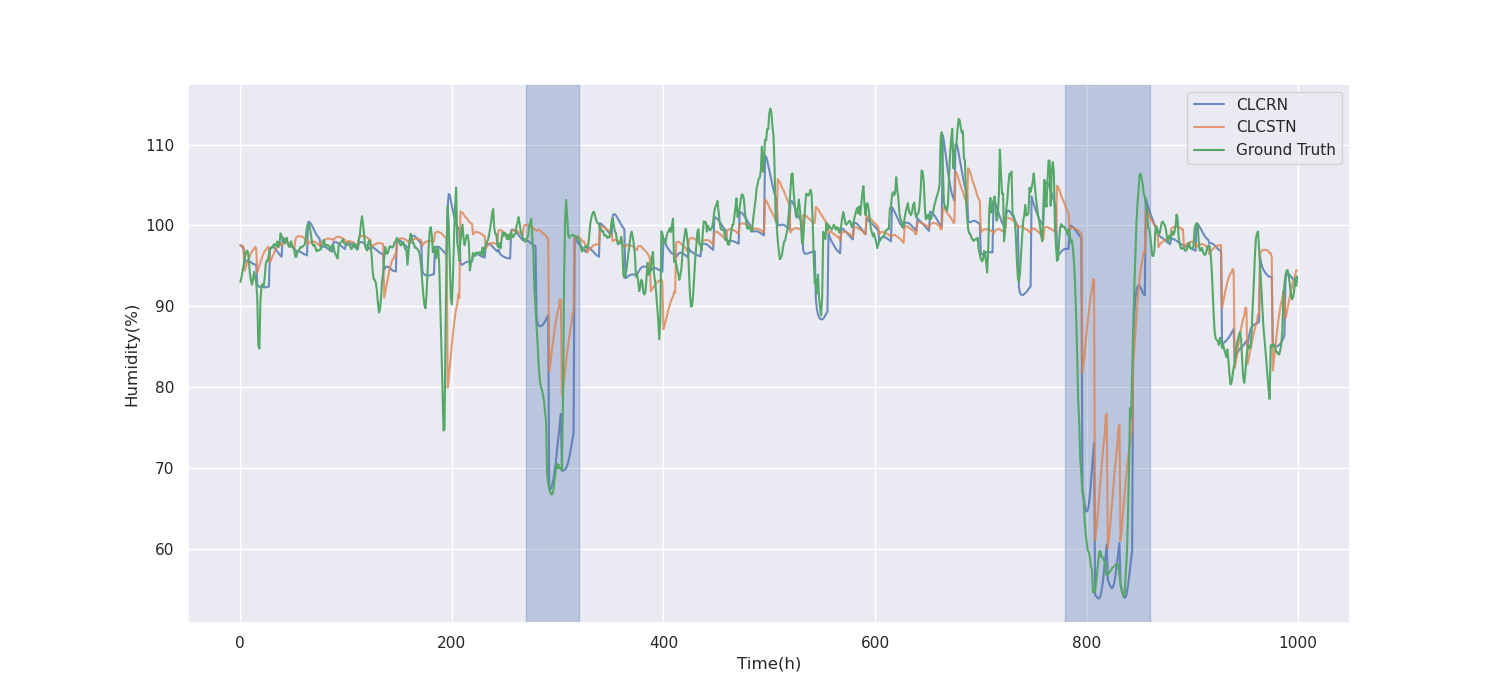}}\\
            \vspace*{-0.2cm}
            \addtocounter{figure}{-1}
\end{figure}
\begin{figure}[H]
    \addtocounter{figure}{1}
    \vspace*{-0.3cm}
    \subfigure[Predictions on Wind on $1000$-th node.]{ 
            \includegraphics[width=1.0\linewidth]{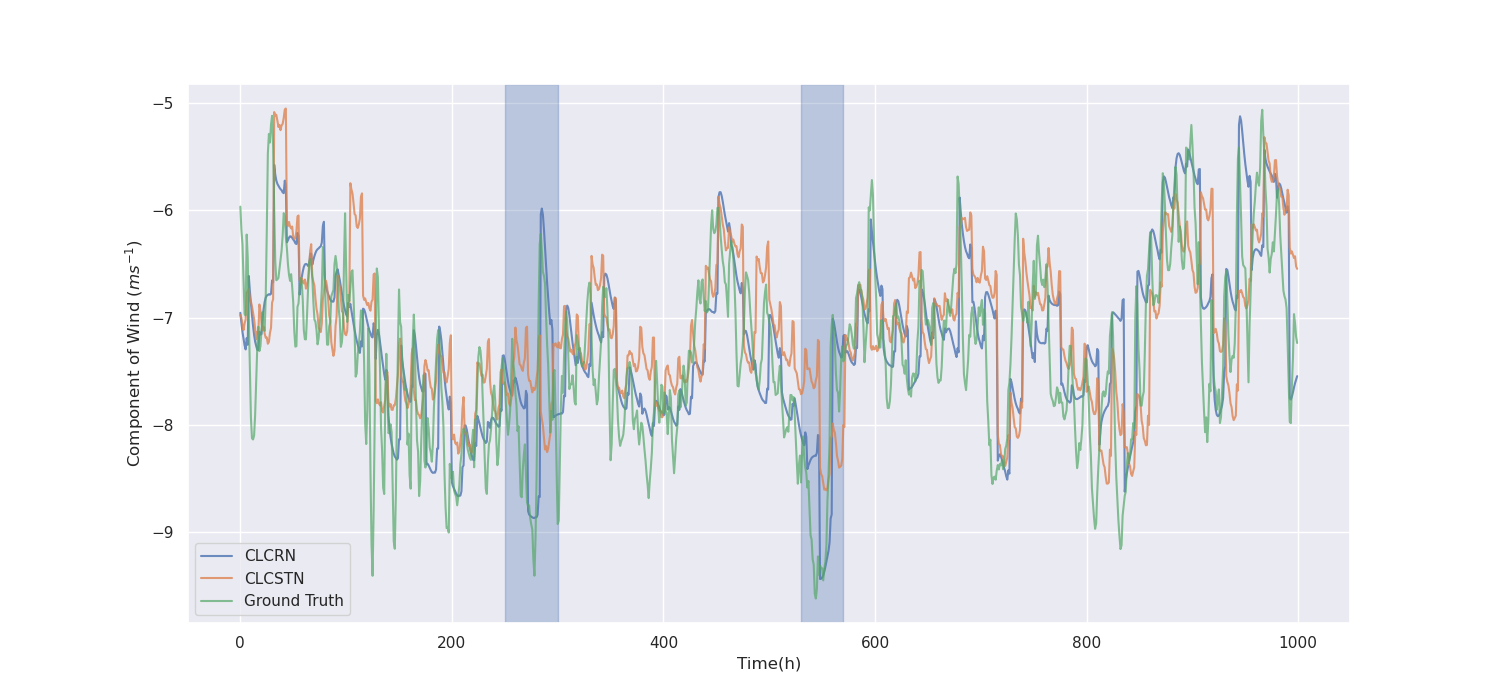}}\\
\subfigure[Predictions on Wind on $2000$-th node.]{ 
    \includegraphics[width=1.0\linewidth]{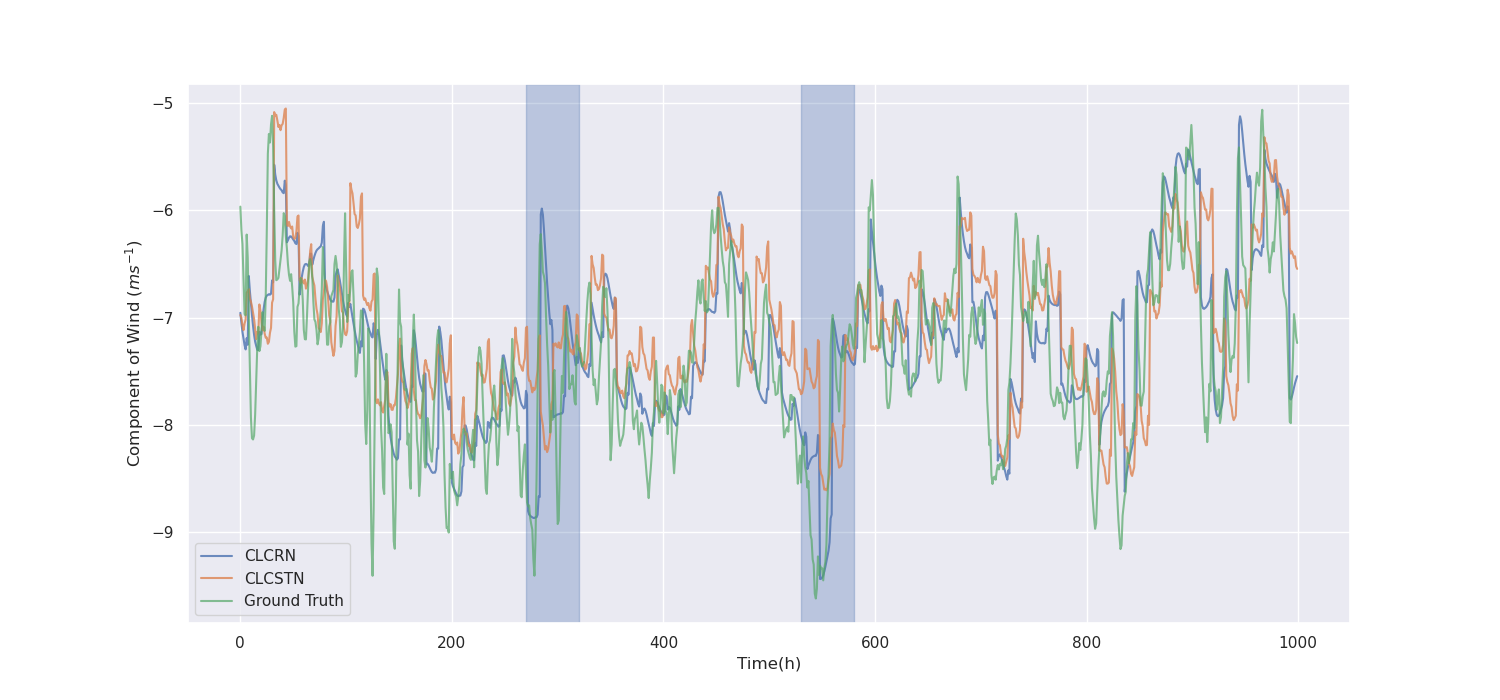}}\\
        \caption{Visualization of two methods on different datasets.} \label{fig:timeseriespred}    
\end{figure} 
It shows that the Temperature dataset is of gentle slope, and dramatic fluctuations and changes are rare, so the prediction generated by CLCSTN is usually good, and metrics are comparably competitive. However, in other dataset, the change is significant and drastic, so CLCRN gets more accurant predition than it. 
\subsection{D4: Sensitivity of hyperparameters}
The sensitivity analysis of parameters in details are given in this part.
\begin{figure}[H]
    \centering
    \subfigure[Sensitivity of model performance on layer number on Temperature.]{ 
        \includegraphics[width=1.0\linewidth]{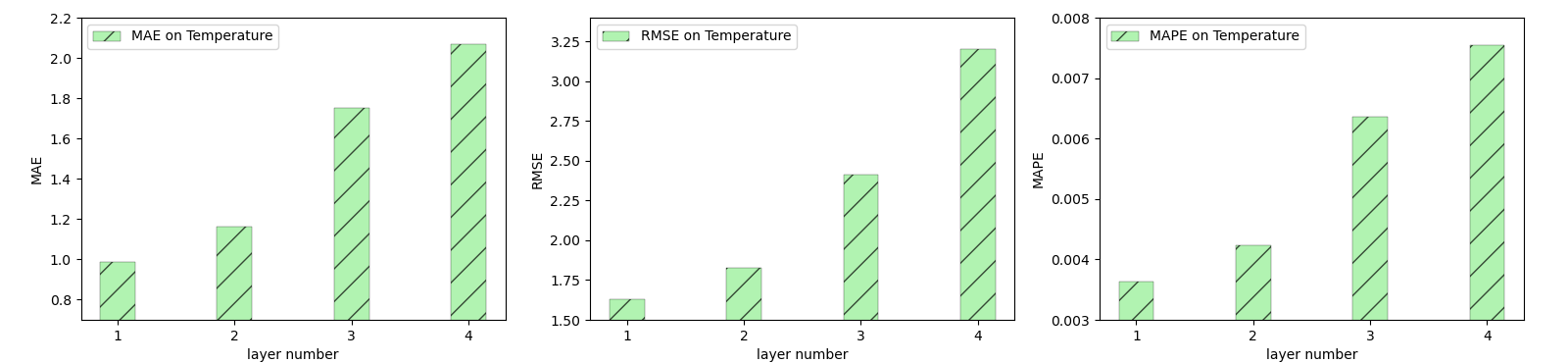}}\\
        \vspace*{-0.2cm}
    \subfigure[Sensitivity of model performance on layer number on Cloud cover.]{ 
        \includegraphics[width=1.0\linewidth]{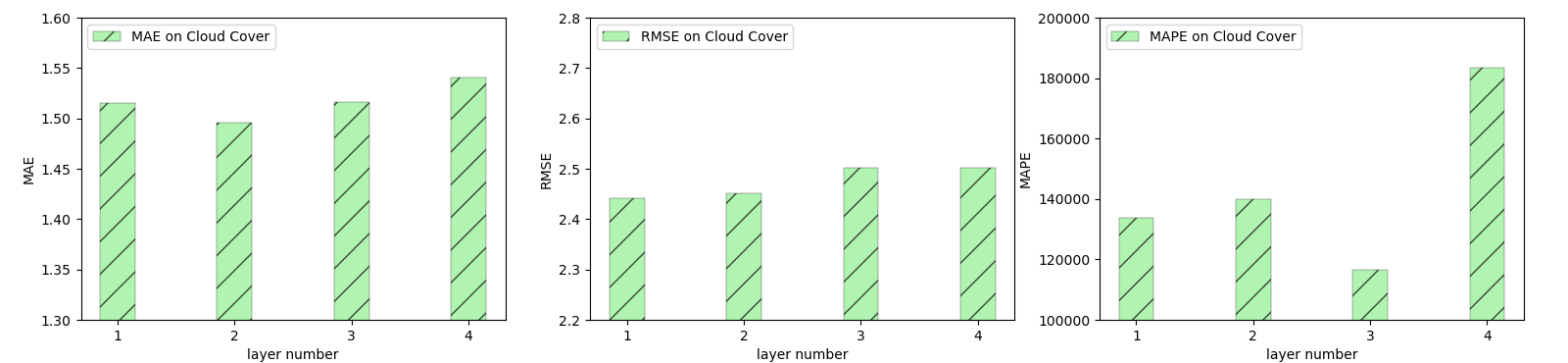}}\\
    \vspace*{-0.2cm}
    \addtocounter{figure}{-1}
\end{figure}
\begin{figure}[H]
    \subfigure[Sensitivity of model performance on layer number on Humidity.]{ 
        \includegraphics[width=1.0\linewidth]{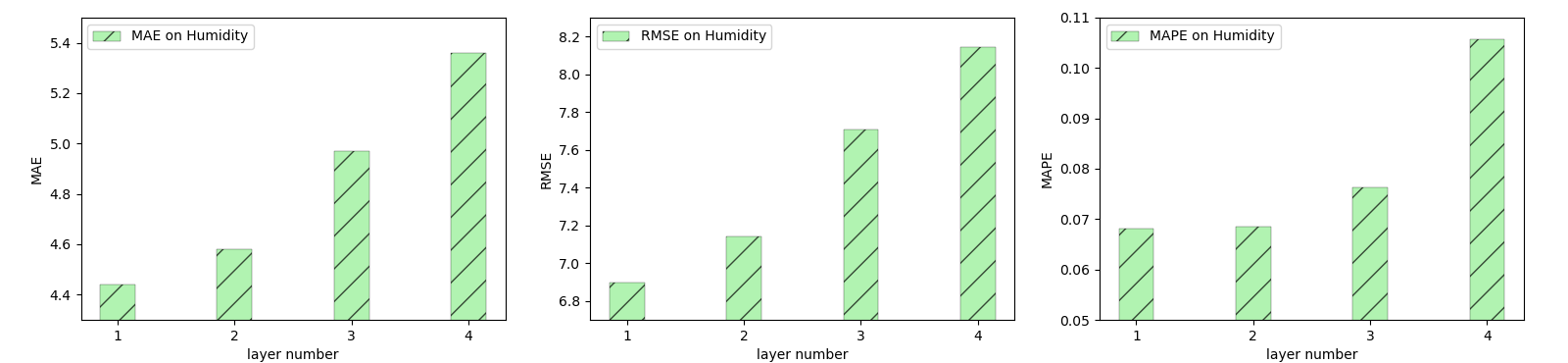}}\\
        \vspace*{-0.2cm}
    \subfigure[Sensitivity of model performance on layer number on Wind.]{ 
        \includegraphics[width=1.0\linewidth]{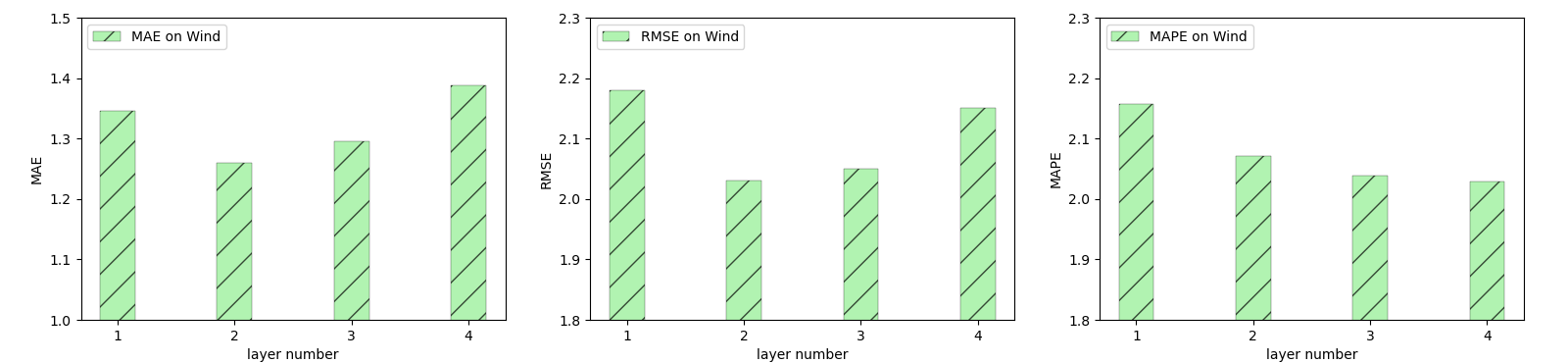}}\\
        \vspace*{-0.2cm}
    \caption{Sensitivity analysis on layer number.}\label{fig:ablationlayernum}
\end{figure}
\paragraph{Layer number.} It can be conclude according to Fig. ~\ref{fig:ablationlayernum} that when the layer number increase, the model parameter increase fast, and the over-fitting affects are more serious. Therefore, we recommend that the layer number is choosen in one or two. When the dataset is relative stationary, the layer number should be one. 

\begin{figure}[ht]
    \centering
    \subfigure[Sensitivity of model performance on neighbor number K on Temperature.]{ 
        \includegraphics[width=1.0\linewidth]{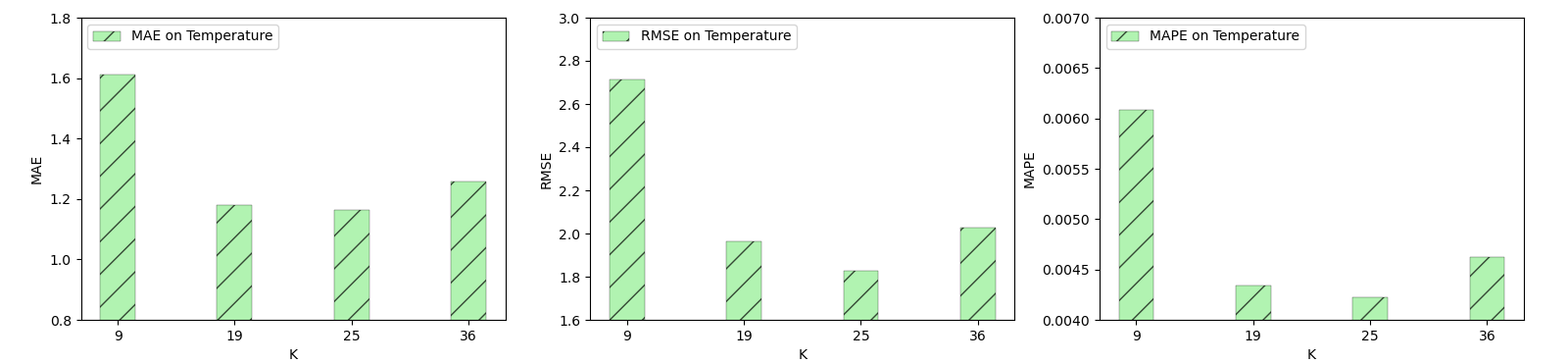}}\\
        \vspace*{-0.2cm}
    \subfigure[Sensitivity of model performance on neighbor number K on Cloud cover.]{ 
        \includegraphics[width=1.0\linewidth]{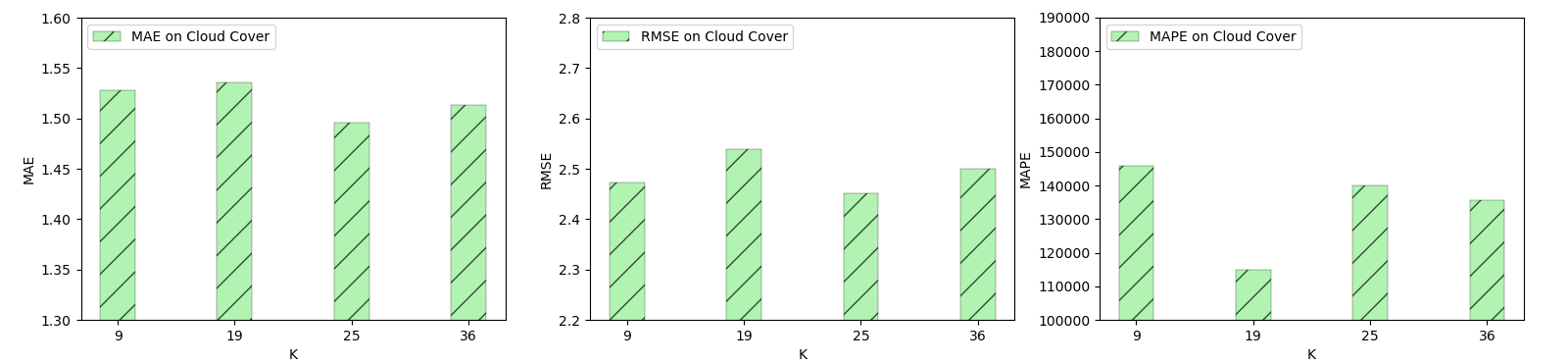}}\\
        \vspace*{-0.2cm}
    \subfigure[Sensitivity of model performance on neighbor number K on Humidity.]{ 
        \includegraphics[width=1.0\linewidth]{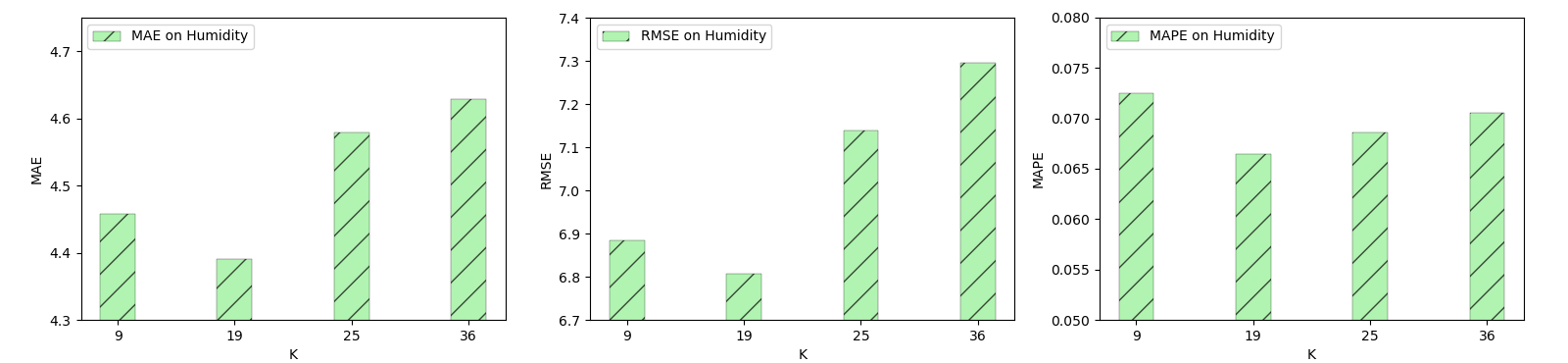}}\\
        \vspace*{-0.2cm}
    \subfigure[Sensitivity of model performance on neighbor number K on Wind.]{ 
        \includegraphics[width=1.0\linewidth]{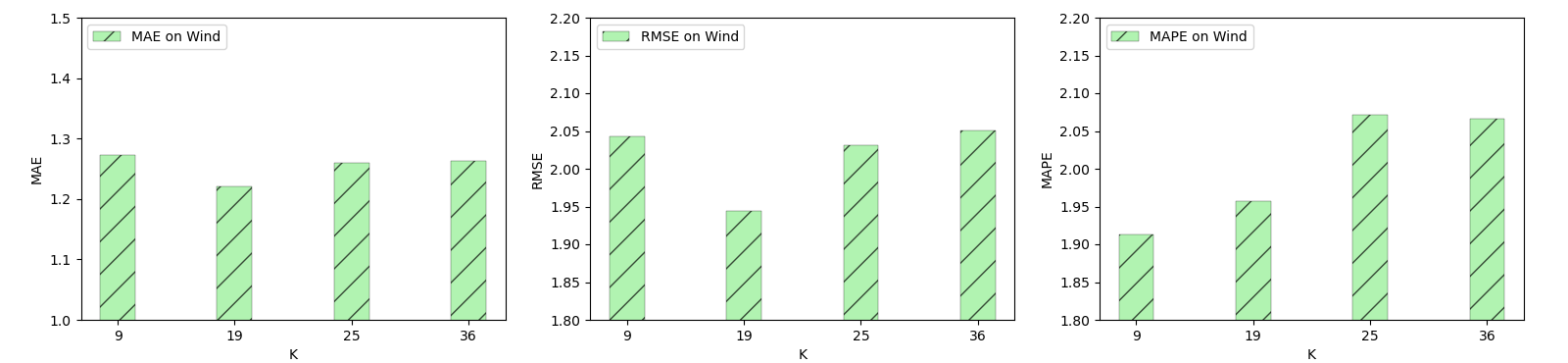}}\\
        \vspace*{-0.2cm}
    \caption{Sensitivity analysis on neighbor number.}\label{fig:ablationK}
\end{figure}

\paragraph{Neighbor Number.} Fig.~\ref{fig:ablationK} shows that the effects of K on model performance. It can be concluded that with the increase of K, CLGRN will perform better, because for each point, it can obtain messages from farther points directly. However, when the K is too large, performance is likely to be compromised due to the aggregation of redundant and irrelevant messages from uncorrelated distant nodes. 

\begin{figure}[ht]
    \centering
    \subfigure[Sensitivity of model performance on hidden units on Temperature.]{ 
        \includegraphics[width=1.0\linewidth]{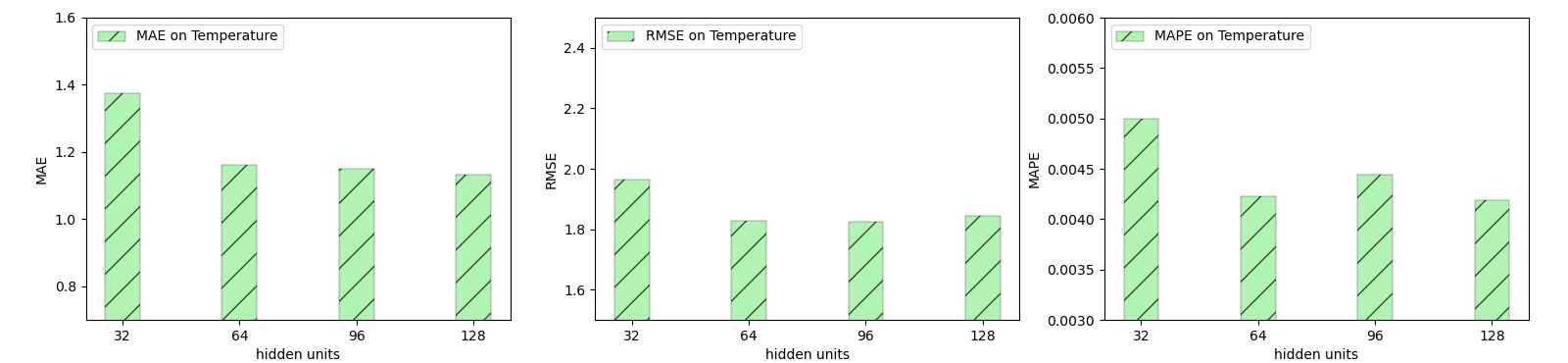}}\\
        \vspace*{-0.2cm}
    \subfigure[Sensitivity of model performance on hidden units on Cloud cover.]{ 
        \includegraphics[width=1.0\linewidth]{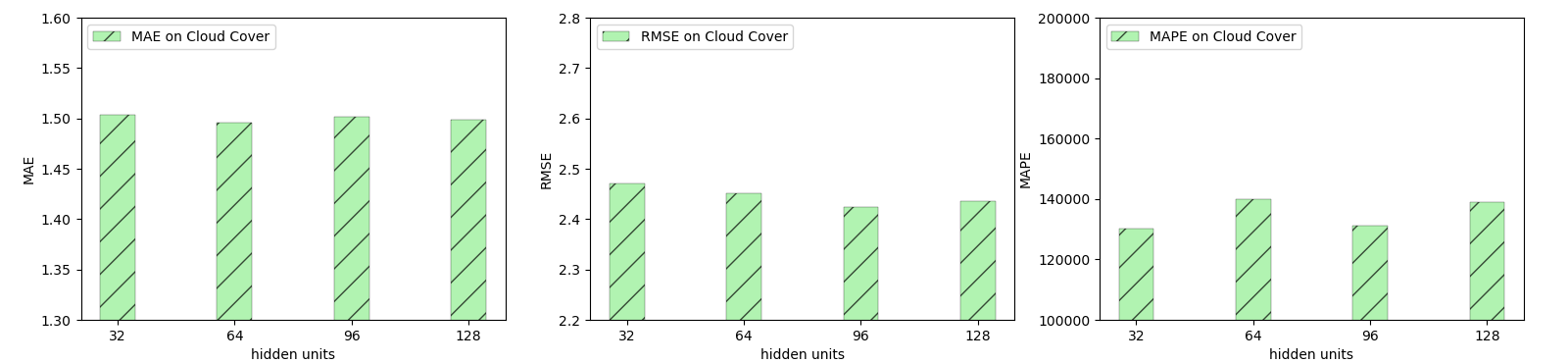}}\\
        \vspace*{-0.2cm}
    \subfigure[Sensitivity of model performance on hidden units on Humidity.]{ 
        \includegraphics[width=1.0\linewidth]{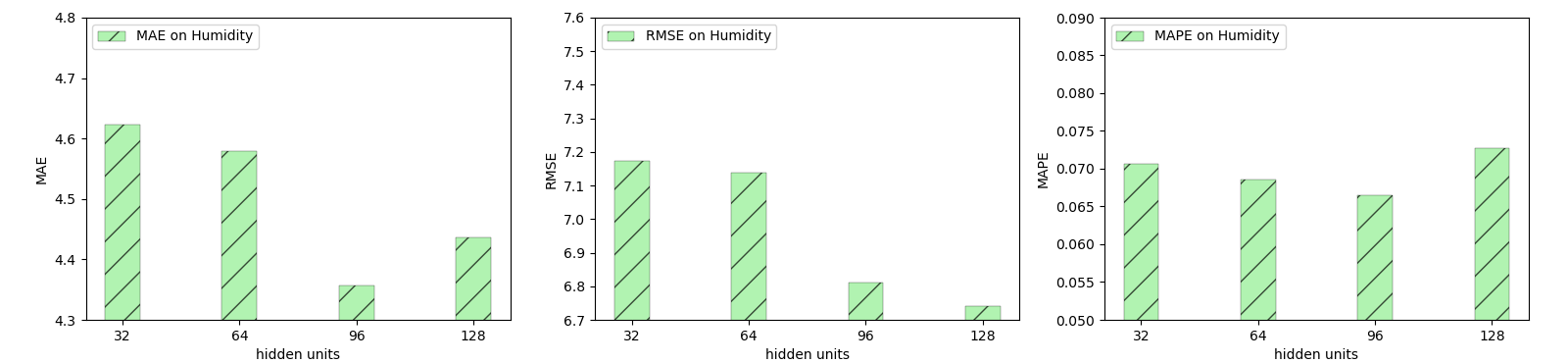}}\\
        \vspace*{-0.2cm}
    \subfigure[Sensitivity of model performance on hidden units on Wind.]{ 
        \includegraphics[width=1.0\linewidth]{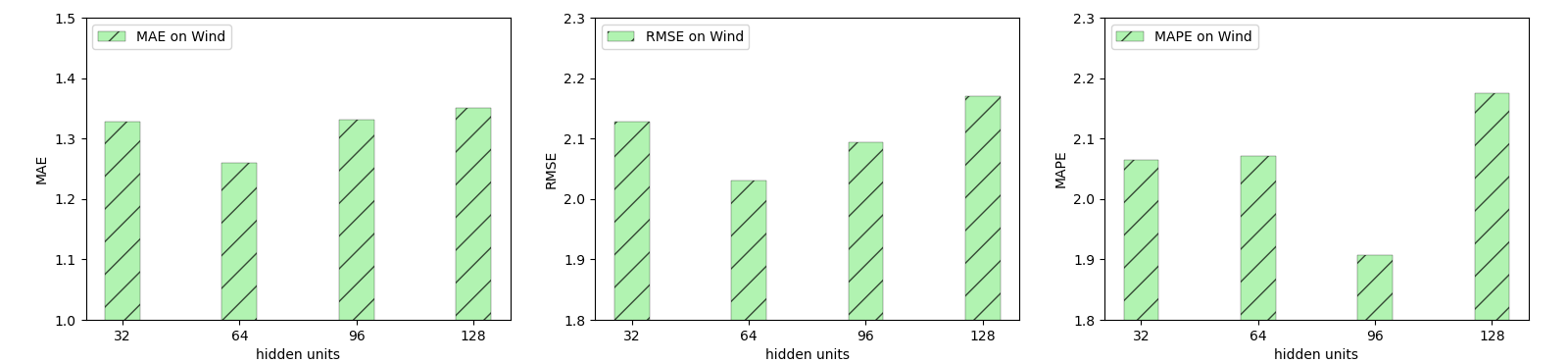}}\\
        \vspace*{-0.2cm}
    \caption{Sensitivity analysis on neighbor number.}\label{fig:ablationhidden}
\end{figure}
\paragraph{Hidden Units.} According to Fig. ~\ref{fig:ablationhidden}, in most cases, the model performs better when the hidden units increase due to the enhanced expressivity of the model. However, the increase of hidden units will dramatically raise the cost of computational time and space. In this way, we recommend that it should be choosen as 64 to 96.

\begin{figure*}[ht]
    \centering
    \subfigure[Comparison on Temperature]{ 
        \includegraphics[width=1\linewidth]{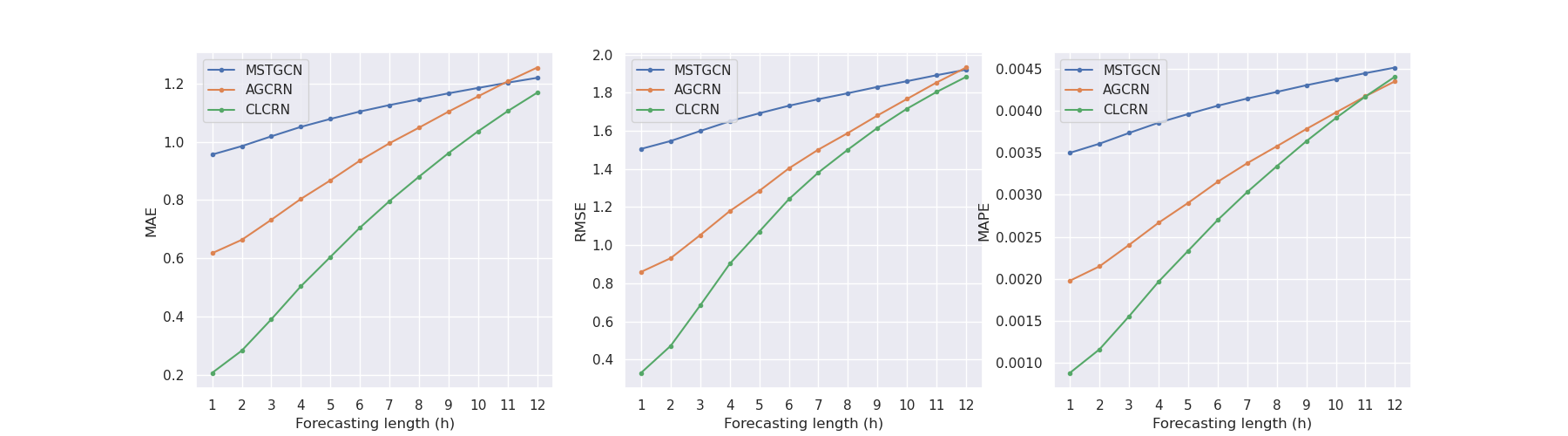}}\\
    \subfigure[Comparison on Cloud cover]{
        \includegraphics[width=1\linewidth]{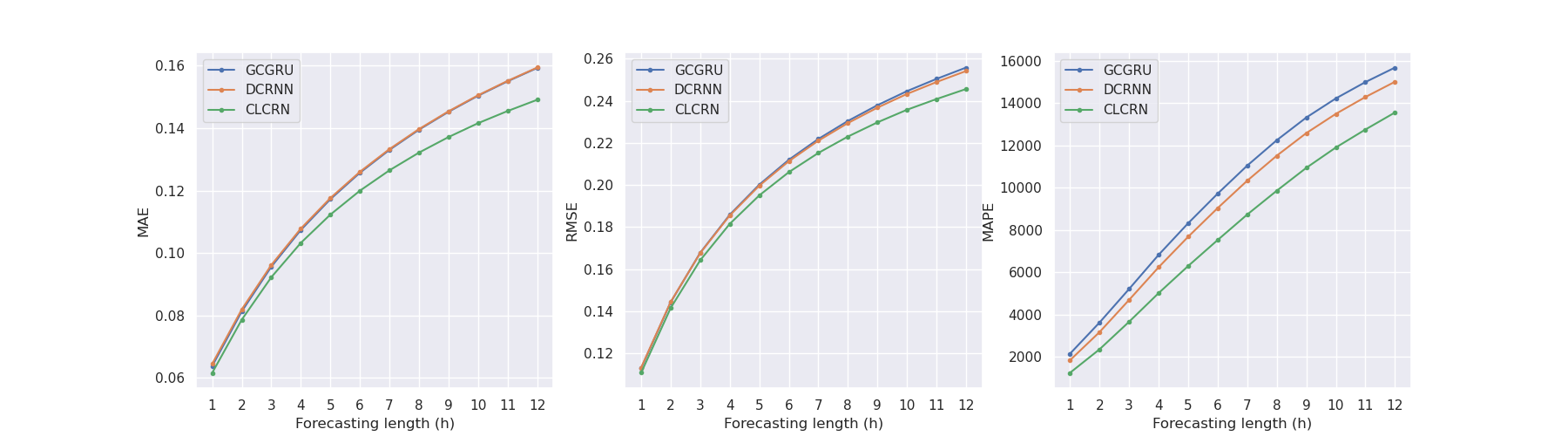}}\\
    \subfigure[Comparison on Humidity]{
        \includegraphics[width=1\linewidth]{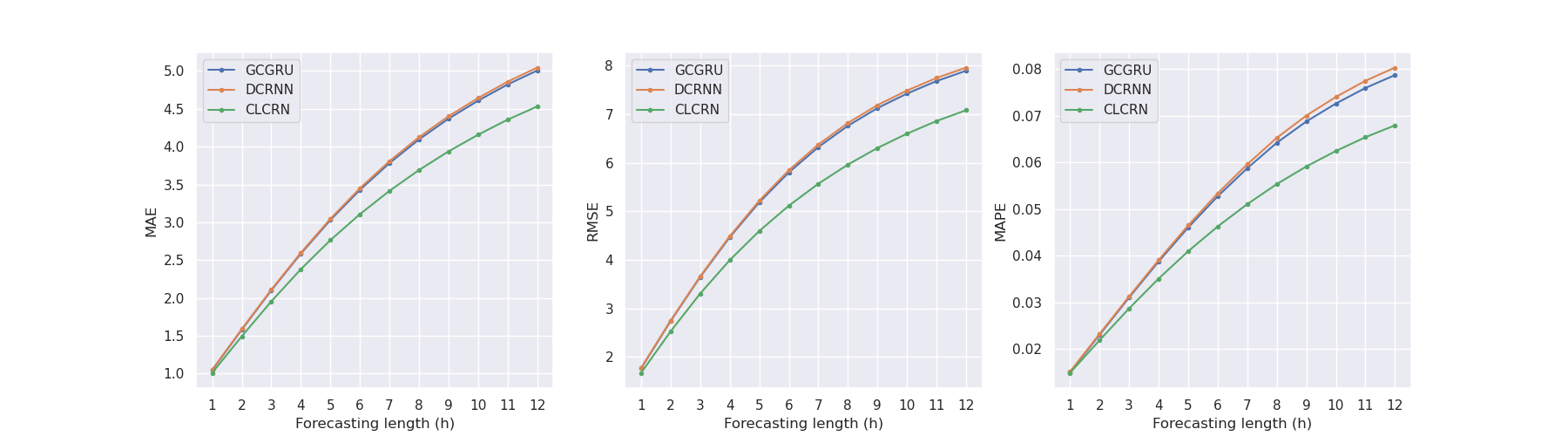}}\\
    \subfigure[Comparison on Wind]{
        \includegraphics[width=1\linewidth]{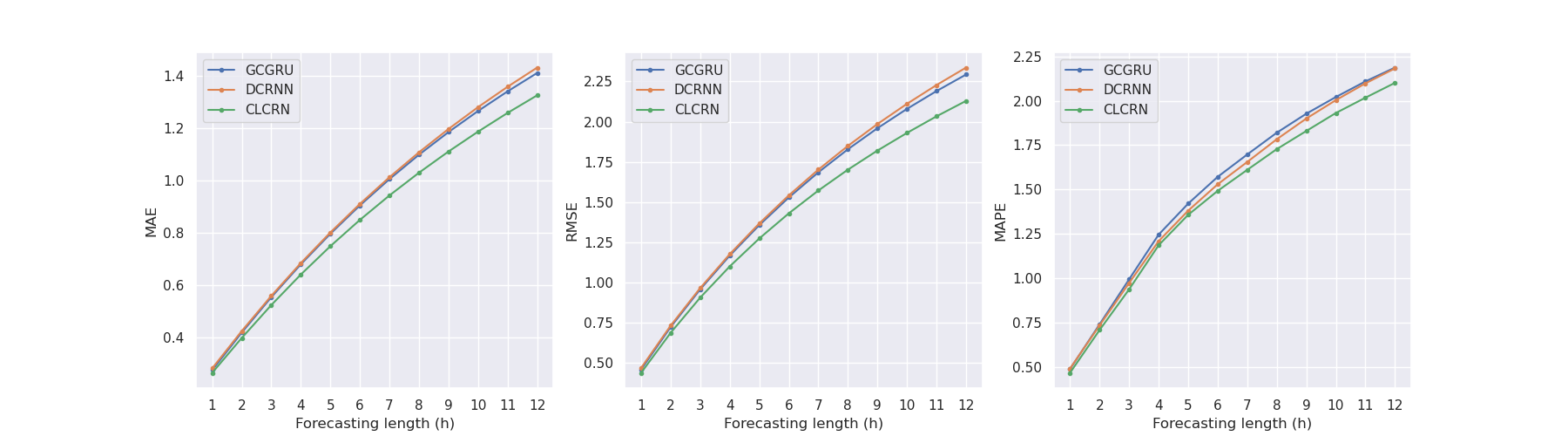}}
        \caption{Overall comparison on four datasets.}
        \label{fig:overallcomp}    
\end{figure*}

\end{document}